%% file: main.tex
\documentclass{article}

\usepackage[final,nonatbib]{neurips_2022}

\usepackage[utf8]{inputenc} 
\usepackage[T1]{fontenc}    
\usepackage{hyperref}       
\usepackage{url}            
\usepackage{booktabs}       
\usepackage{amsfonts}       
\usepackage{nicefrac}       
\usepackage{microtype}      
\usepackage{xcolor}         
\usepackage{multirow}

\usepackage{graphicx}
\usepackage{wrapfig}
\usepackage{floatrow}
\usepackage{subcaption}
\usepackage[title,toc,titletoc,page]{appendix}

\hypersetup{
    colorlinks=true,
    linkcolor=red,
    citecolor=cyan,
    filecolor=magenta,      
    urlcolor=cyan,
    }

\usepackage{amsmath,amssymb}
\usepackage{textgreek}
\usepackage{algorithm}
\usepackage{algpseudocode}

\newfloatcommand{figurebox}{figure}[\nocapbeside][\dimexpr(\textwidth-\columnsep)/2\relax]
\newfloatcommand{tablebox}{table}[\nocapbeside][\dimexpr(\textwidth-\columnsep)/2\relax]
\newfloatcommand{figureboxl}{figure}[\nocapbeside][0.95\dimexpr(\textwidth-\columnsep)/2\relax]
\newfloatcommand{tableboxl}{table}[\nocapbeside][1.05\dimexpr(\textwidth-\columnsep)/2\relax]
\newfloatcommand{figureboxf}{figure}[\nocapbeside][2\dimexpr(\textwidth-\columnsep)/2\relax]
\newfloatcommand{tableboxf}{table}[\nocapbeside][2\dimexpr(\textwidth-\columnsep)/2\relax]

\newcommand{\etal}{\textit{et al}.}

\newcommand{\lsh}{\textcolor{black}}
\newcommand{\xw}{\textcolor{black}}

\title{Dataset Distillation via Factorization}

\author{Songhua Liu \quad
 Kai Wang \quad Xingyi Yang \quad Jingwen Ye  \quad  Xinchao Wang \\
 National University of Singapore \\
 \texttt{\{songhua.liu,e0823044,xyang\}@u.nus.edu, \{jingweny,xinchao\}@nus.edu.sg}
}

\begin{document}

\maketitle

\begin{abstract}
  In this paper, we study \xw{dataset distillation (DD)}, from a novel perspective and introduce a \emph{dataset factorization} approach, termed \emph{HaBa}, which is a plug-and-play strategy portable to any existing DD baseline. 
  Unlike conventional DD approaches that aim to produce distilled and representative samples, \emph{HaBa} explores decomposing a dataset into two components: data \emph{Ha}llucination networks and \emph{Ba}ses, where the latter is fed into the former to reconstruct image samples. 
  The flexible combinations between bases and hallucination networks, therefore, equip the distilled data with exponential informativeness gain, which largely increase the representation capability of distilled datasets. 
  To furthermore increase the data efficiency of compression results, we further introduce a pair of adversarial contrastive \xw{constraints} on the resultant hallucination networks and bases, which increase the diversity of generated images and inject more discriminant information into the factorization.
  Extensive comparisons and experiments demonstrate that our method can yield significant improvement on downstream classification tasks compared with previous state of the arts, while reducing the total number of compressed parameters by up to 65\%. 
  Moreover, distilled datasets by our approach also achieve \textasciitilde10\% higher accuracy than baseline methods in cross-architecture generalization. 
  Our code is available \href{https://github.com/Huage001/DatasetFactorization}{here}.
\end{abstract}

\section{Introduction}

The success of deep models on a variety of vision tasks, such as image classification~\cite{krizhevsky2012imagenet,dosovitskiy2020image,SuchengCVPR22}, object detection~\cite{ren2015faster,redmon2018yolov3}, and semantic segmentation~\cite{shelhamer2017fully,xie2021segformer,HuihuiAAAI21}, 
is largely attributed to the huge amount of data used for training and various pre-trained models~\cite{XingyiECCV22}.
However, the sheer amount of data introduces significant obstacles for storage, transmission, and data pre-processing.
Besides, publishing raw data
inevitably brings about privacy or 
copyright issue in practice~\cite{shokri2015privacy,dong2022privacy}. 
\xw{To alleviate these problems, Wang \etal~\cite{wang2018dataset} pioneer the research of dataset distillation (DD), to distill a large dataset into a synthetic one with only a limited number of samples, so that the training efforts with the distilled dataset for downstream models on the original dataset can be largely reduced, which facilitates a series of applications like continual learning~\cite{sangermano2022sample,rosasco2021distilled,wiewel2021soft,masarczyk2020reducing} and black-box optimization~\cite{chen2022bidirectional}.} 
\xw{Due the significant practical value of DD, 
many endeavours have been made on this area~\cite{zhao2020dataset,zhao2021dataset,zhao2021distribution,wang2022cafe,kim2022dataset,cazenavette2022dataset,zhou2022dataset}
to design novel supervision signals to train the synthetic datasets and to further improve their  
performances. }

\begin{figure}[!t]
\begin{floatrow}[1]
\figureboxf{}{
  \centering
  \includegraphics[width=0.9\textwidth]{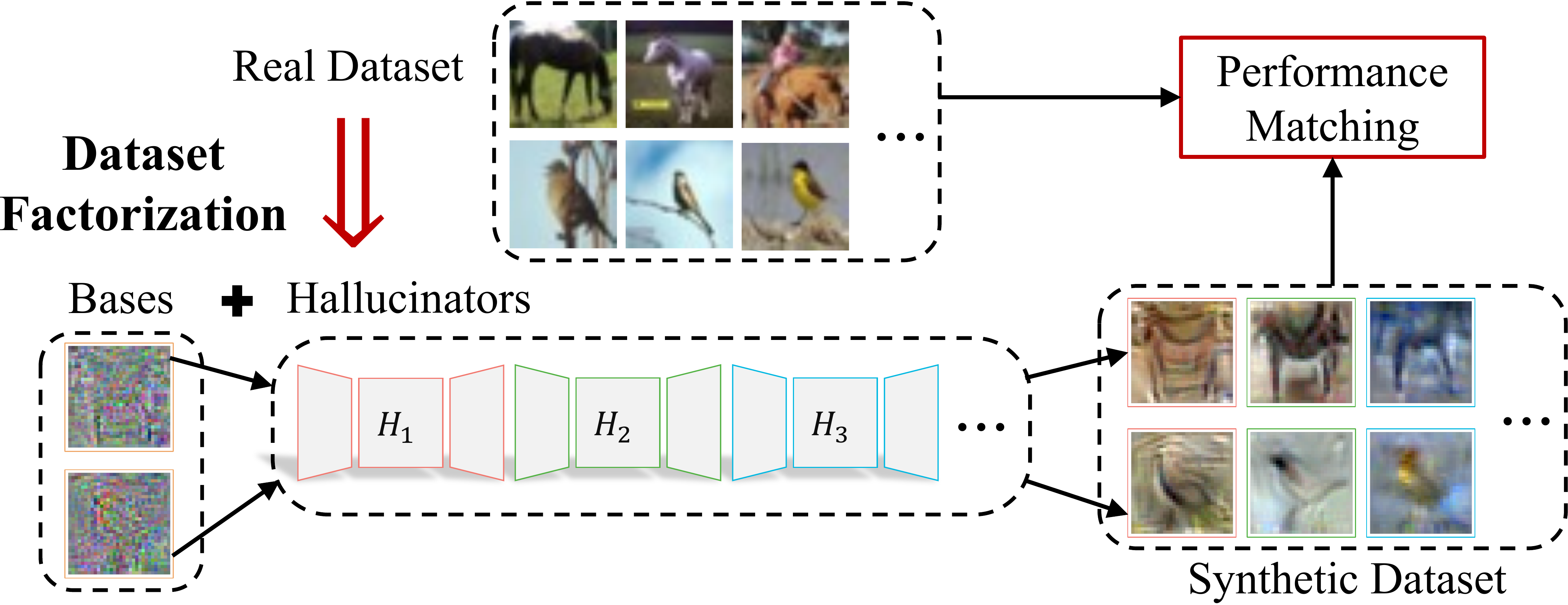}
  \caption{Intuition of our hallucinator-basis factorization for dataset distillation.}
  \label{fig:motivation}
}
\end{floatrow}
\vspace{-0.6cm}
\end{figure}

Nevertheless, there is a potential drawback in conventional settings of DD: 
it largely treats each synthetic sample independently and ignores the inter coherence and relationship between different instances. 
As such, the information embraced by each sample,
despite distilled, 
is by nature limited.
Using the synthetic samples for training downstream
models, therefore, 
inevitably leads to the loss of dataset information.
Moreover, the few distilled samples are incompatible with the enormous number of parameters in a deep model and may yield the risk of overfitting. 

\begin{wrapfigure}{r}{4cm}
\begin{floatrow}[1]
\figureboxf{}{
  \centering
  \vspace{-1em}
  \includegraphics[width=\linewidth]{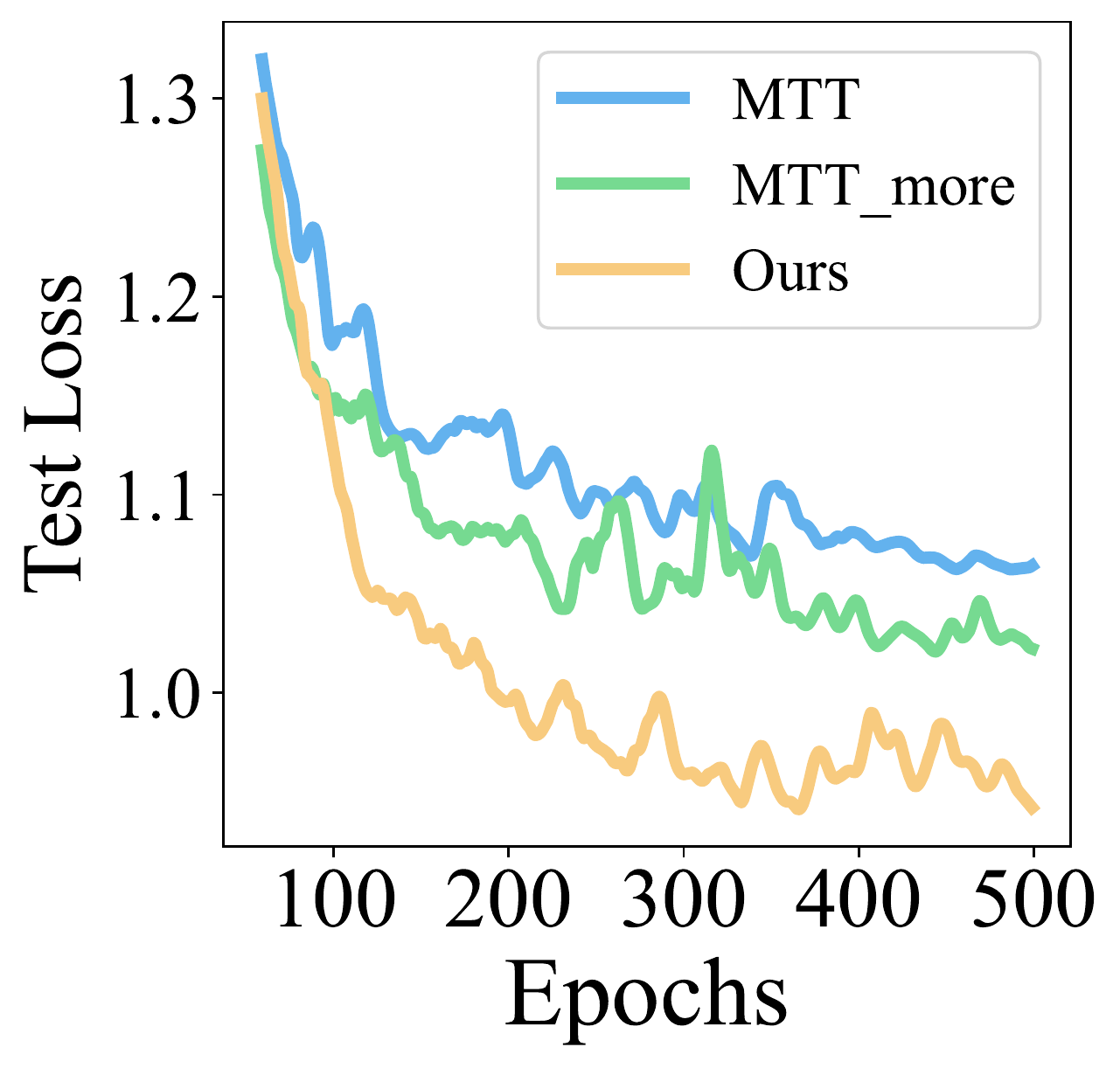}
  \vspace{-2.0em}
  \caption{Visualization of test loss using synthetic datasets generated by MTT, MTT with multiple checkpoints, and ours.}
  \label{fig:loss}
  }
  \end{floatrow}
  \vspace{-1em}
  \end{wrapfigure}

To verify these potential issues, we conduct a pre-experiment on CIFAR10 dataset with 10 synthetic images per class, using MTT~\cite{cazenavette2022dataset}, the current SOTA solution on DD, as the baseline. 
In addition to the baseline setting, we also incorporate all the checkpoint synthetic datasets after each 100 DD iterations in the convergent stage to train the downstream model. 
\xw{Since the synthetic images are fine-tuned during this stage, multiple checkpoints can be viewed as related but different, which may somehow increase the diversity.} 
As a result, it yields overall lower test loss and hence better final results in downstream training, as shown in the blue and green curves in Fig. \ref{fig:loss}, which indicates that \xw{current DD solutions can be potentially improved by leveraging some sample-wise relationships to diversify the distilled data}. 
Nevertheless, simply involving more data samples may also increase the memory overhead. 
This fact motivates us to ask: \emph{is it possible to encode some shared relationships in a dataset implicitly, instead of storing samples directly, to avoid such additional storage costs?}

We show in this paper that, it can indeed be made possible
through reformulating the DD task as a  \emph{factorization} problem.
As shown in Fig. \ref{fig:motivation}, we propose a novel perspective dubbed \emph{HaBa}, to factorize a dataset into two compositions: data \emph{Ha}llucination networks and \emph{Ba}ses. 
A data hallucination network, or hallucinator, can take any basis as input and output the corresponding hallucinated image. 
Supervised by the training objective of DD, a set of hallucinators can synthesize multiple samples from a common basis and are optimized to extract effective relationships among different samples in original datasets explicitly. 
\xw{In this way, information of $|\mathcal{H}|\times |\mathcal{B}|$ images can be included for a factorization result with $|\mathcal{H}|$ hallucinators and $|\mathcal{B}|$ bases via arbitrary pair-wise combination, which improves the data efficiency of traditional DD exponentially. 
As shown in the yellow curve in Fig. \ref{fig:loss}, with the same budget on the storage, our strategy achieves better test performance compared with the MTT baseline}. 

To further increase the informativeness of factorized results, we introduce a pair of adversarial contrastive \xw{constraints} to promote sample-wise diversity. 
The goal of \lsh{HaBa} is to minimize the \lsh{correlation} among images composed of different hallucinators but a common basis, while an adversary tries to maximize it. 
Such an adversarial scheme, in turn, enforces the hallucinators to produce diversified images and \xw{increases the amount of useful information}. 

Notably, HaBa is a versatile strategy that can be built upon existing DD baselines, since it is compatible with any training objective 
for measuring the similarity between downstream performances
as shown in Fig. \ref{fig:motivation}, 
We conduct extensive experiments to demonstrate the advantages of the proposed method over baseline ones. 
In all benchmarks and comparisons, HaBa produces significant and consistent improvement on training downstream models, while reducing the total number of compressed parameters by up to 65\%. 
Furthermore, it demonstrates strong cross-architecture generalization ability with accuracy improvement higher than 10\%. 
Our contributions are summarized as follows:
\begin{itemize}
    \item We study dataset factorization, a 
    novel perspective to explore dataset distillation,
    and propose a novel approach termed HaBa
    for hallucinator-basis factorization. 
    \item We present a pair of adversarial contrastive objectives to further increase the data diversity and information capability. 
    \item HaBa is a plug-and-play scheme compatible with all existing training objectives of DD and can yield significant and consistent improvement over the state of the arts. 
\end{itemize}

\section{Related Works}

The goal of \xw{dataset distillation (DD)} is to optimize a smaller synthetic dataset such that it is capable to take place of original one for training downstream tasks, which is different from coreset selection~\cite{bachem2017practical,chen2012super,har2007smaller,sener2017active,tsang2005core}, another branch for dataset compression, directly selecting samples from raw datasets. 
In this section, we provide a detailed review of previous methods in DD. 

Motivated from knowledge distillation~\cite{hinton2015distilling,gou2021knowledge,YidingCVPR20,YidingNIPS20} aiming at model compression, Wang~\etal~\cite{wang2018dataset} introduce the concept of dataset distillation for dataset compression. 
The idea is to optimize the synthetic images so that they can minimize loss functions of downstream tasks, where a bilevel optimization algorithm~\cite{finn2017model} is involved. 
Following this routine, several works further consider learnable labels beyond samples~\cite{bohdal2020flexible,sucholutsky2021soft}. 
Subsequently, Zhao~\etal~\cite{zhao2020dataset} and several following approaches~\cite{zhao2021dataset,lee2022dataset} consider matching gradients of a downstream model produced by synthetic samples and real images, \xw{which improve the performance significantly}. 
Most recently, Cazenavette \etal~\cite{cazenavette2022dataset} argue that single-iteration gradient matching may lead to inferior performance due to error accumulation across multiple steps and thereby propose to match long-range training dynamics of an expert trained on the original dataset. 
As an alternative method to profile training effects produced by different sets,  Nguyen~\etal~\cite{nguyen2020dataset,nguyen2021dataset} also introduce the kernel ridge-regression approach based on the Neural Tangent Kernel (NTK) in infinitely wide convolutional networks~\cite{jacot2018neural}. 

Apart from matching training effects, there are also methods matching data distributions between original and synthetic datasets. 
For instance, Zhao~\etal~\cite{zhao2021distribution} propose a simple but effective Maximum Mean Discrepancy (MMD) \xw{constraint} for DD, which does not involve the training of downstream models and enjoys superior training efficiency. 
Wang~\etal~\cite{wang2022cafe} propose CAFE, explicitly attempting to align the synthetic and real distributions in the feature space of a downstream network. 

Above mentioned methods are dedicated to exploring suitable training objectives and pipelines for DD. 
However, there are few works concerning improving the data efficiency for distilled samples. 
Although Zhao~\etal~\cite{zhao2021dataset} propose differentiable siamese augmentation (DSA) to enrich the training data, the augmentation operations used, \textit{e.g.}, crop, flip, scale, and rotation, cannot encode any information about the target datasets. 
In this paper, we study the task in a factorization perspective, to factorize a dataset into two different compositions: data hallucination networks and bases. 
Both parts carry important knowledge of the raw dataset. 
For downstream training, hallucinators and bases can perform arbitrary pair-wise combination, \textit{i.e.}, sending any basis to any hallucinator, to create a training sample. 
The idea of factorization can improve the diversity of distilled training datasets significantly, without introducing additional costs for storage. 
It is also a versatile strategy compatible with all aforementioned DD methods, which will be demonstrated in the experiment part. 

\textbf{Concurrent Works on Efficient Distilled Dataset Parameterization:} 
As a concurrent work, Kim \etal~\cite{kim2022dataset} propose IDC for efficient synthetic data parameterization. 
It reveals that only storing down-sample version of synthetic images and conducting bilinear upsampling in downstream training would not hurt the performance much. 
Thus, given the same budget of storage, it can store $4\times$ number of $2\times$ down-sample synthetic images compared with the baseline. 
Both IDC and HaBa in this paper are dedicated to improving the data efficiency of synthetic parameters. 
Interestingly, according to the definition of our hallucinator-basis factorization, IDC can in fact be treated as a special case of HaBa, where the hallucinator is a parameter-free upsampling function and each basis has a smaller spatial size. 
Nevertheless, the main focuses for IDC and HaBa are different and they are in fact two orthogonal techniques, which can readily join force to enhance the baseline performance, as discussed in Sec. \ref{sec:4-2}.

\section{Methods}

In this section, we elaborate our proposed method \emph{HaBa} for dataset distillation (DD). 
Assume that there is an original dataset $\mathcal{T}=\{(x_i,y_i)\}_{i=1}^{|\mathcal{T}|}$ with $|\mathcal{T}|$ pairs of a training sample $x_i$ and the corresponding label $y_i$. 
DD targets a synthetic dataset $\mathcal{S}=\{(\hat{x}_i,\hat{y}_i)\}_{i=1}^{|\mathcal{S}|}$ with $|\mathcal{S}|\ll|\mathcal{T}|$ and expects that a model trained on $\mathcal{S}$ can have similar performance than that trained on $\mathcal{T}$. 

Traditional DD methods treat each synthetic sample independently and ignore the inner relationship between different samples within a dataset, which results in poor data/information efficiency. 
Focusing on such drawback, we study DD from a novel perspective and redefine it as a hallucinator-basis factorization problem:
\begin{equation}
    \mathcal{S}=\{H_{\theta_j}\}_{j=1}^{|\mathcal{H}|}\cup\{(\hat{x}_i,\hat{y}_i)\}_{i=1}^{|\mathcal{B}|},
\end{equation}
where there are $|\mathcal{H}|$ hallucination networks and $|\mathcal{B}|$ bases. 
The $j$-th hallucinator is parameterized by $\theta_i$ and we denote it by $H_{\theta_i}$ for $1\leq j\leq |\mathcal{H}|$. 
For downstream training, a training data pair $(\tilde{x}_{ij},\tilde{y}_{ij})$ is created online via sending the $i$-th basis, with any $1\leq i\leq |\mathcal{B}|$, to the $j$-th hallucinator, with any $1\leq j\leq |\mathcal{H}|$, \textit{i.e.}, $\tilde{x}_{ij}=H_{\theta_j}(\hat{x}_i)$. 
In this paper, the label $\tilde{y}_{ij}$ is simply taken as $\hat{y}_i$. 

An overview of our method is shown in Fig. \ref{fig:pipeline}(Left). 
To go deeper into the technical details, we first start with the introduction of our basis and data hallucination network in Sec. \ref{subsec:1}. 
Then, we propose an adversarial contrastive \xw{constraint} to increase data diversity in Sec. \ref{subsec:2}. 
Finally, we present the whole training pipeline of the hallucinator-basis factorization for DD in Sec. \ref{subsec:3}.  

\subsection{Basis and Hallucinator}\label{subsec:1}

\textbf{Basis:} 
Typically, for an image classification dataset $\mathcal{T}=\{(x_i,y_i)\}_{i=1}^{|\mathcal{T}|}$, $x_i\in\mathbb{R}^{h\times w\times c}$ and $y_i\in\{0,1,...,C-1\}$ for each $1\leq i\leq |\mathcal{T}|$, where each $x_i$ is a $c$-channel image with a resolution of $h\times w$, and $C$ is the total number of classes. 
In previous DD methods, the format/shape of synthetic data pairs $(\hat{x},\hat{y})$ has to be held the same as that of real data, so as to make sure the consistency between input and output formats in the training and test time for downstream models. 
By contrast, since hallucinator networks are capable of spatial-wise and channel-wise transformation, the shape of each $\hat{x}_i$, $1\leq i\leq |\mathcal{B}|$, denoted as $h'\times w'\times c'$, is not necessarily the same as that of original samples and thus more flexible. 
And for a classification problem, we do not modify its label space in this paper for simplicity and maintain the categorical format. 

\begin{figure}[!t]
\begin{floatrow}[1]
\figureboxf{}{
  \centering
  \includegraphics[width=\linewidth]{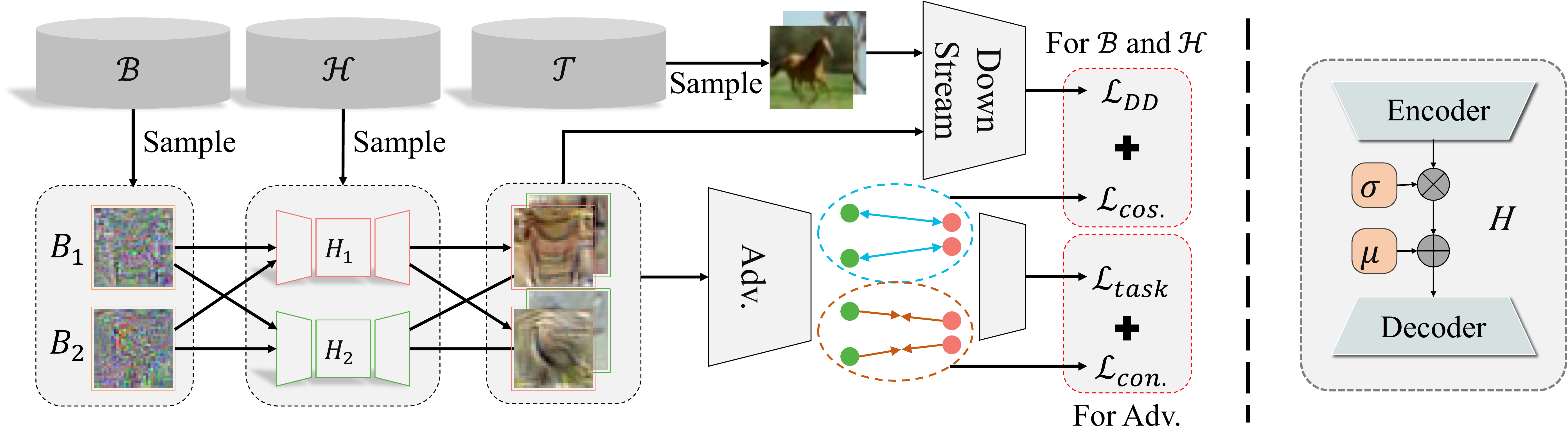}
  \vspace{-0.6cm}
  \caption{Left: Overall pipeline of the proposed hallucinator-basis factorization. $\mathcal{B}$, $\mathcal{H}$, and $\mathcal{T}$ denote sets of bases, hallucinators, and original data respectively. \textit{Adv.} denotes an adversary model. We adopt batch size $2$ here for clarity; Right: Architecture of a hallucinator in detail.}
  \label{fig:pipeline}
  }
  \end{floatrow}
  \vspace{-0.6cm}
\end{figure}

\textbf{Hallucinator:} 
Given a basis $\hat{x}\in\mathbb{R}^{h' \times w' \times c'}$, a data hallucination network, aims to create a new image $\tilde{x}\in\mathbb{R}^{h\times w\times c}$ based on $\hat{x}$, which can be viewed as a conditional image generation problem. 
Inspired by image style transfer~\cite{Jing_2018_ECCV,huang2017arbitrary,jing2020dynamic,SonghuaECCV22}, a typical conditional image generation problem, we devise an encoder-transformation-decoder based architecture for hallucinators, as shown in Fig. \ref{fig:pipeline}(Right). 
Specifically, the encoder, denoted as $enc$, is composed of CNN blocks, which non-linearly maps an input $\hat{x}$ to a feature space $\mathbb{R}^{h''\times w''\times c''}$. 
Then, an affine transformation with scale $\sigma$ and shift $\mu$ is conducted on the derived feature, where $\sigma$ and $\mu$ are treated as network parameters in this paper. 
At last, the decoder $dec$ under a symmetric CNN architecture with $enc$ projects the transformed feature back to the image space. 
Formally, this process can be written as:
\begin{equation}
    \hat{f}=enc(\hat{x}),\hspace{3mm} \tilde{f}=\sigma\times\hat{f}+\mu,\hspace{3mm} \tilde{x}=dec(\tilde{f}), 
\end{equation}
where the multiplication is element-wise operation. 
There are $|\mathcal{H}|$ hallucinators in the whole factorization pipeline and each would be trained to implicitly encode some sample-wise relations by its network parameters. 

\subsection{Adversarial Contrastive Constraint}\label{subsec:2}

Ideally, the knowledge encoded by different hallucinators should be as different/orthogonal as possible to get the most benefits for each individual. 
To instantiate such regularization, let's consider two composed images $\tilde{x}_{ij}$ and $\tilde{x}_{ik}$ from two different hallucinators $H_{\theta_j}$ and $H_{\theta_k}$ but a common basis $\hat{x}_i$. 
The divergence between $\tilde{x}_{ij}$ and $\tilde{x}_{ik}$ is expected to be large. 
To measure the divergence, a feature extractor is required to map an input image to a feature space, and how to train such a feature extractor to find an appropriate feature space is of great importance. 

In this paper, we formalize the training of hallucinators and the feature extractor as a min-max game in a self-consistent manner, where the feature extractor desires to minimize the divergence between $\tilde{x}_{ij}$ and $\tilde{x}_{ik}$ while hallucinators, as well as bases, are optimized to maximize it so that the two players can reinforce each other. 
In specific, the feature extractor, denoted as $F$ and parameterized by $\psi$, is typically a CNN structure for the downstream task and we adopt features at the last hidden layer before the output layer, denoted as $F_{-1}(\tilde{x}_{ij})$ and $F_{-1}(\tilde{x}_{ik})$. 
$F$ is optimized to maximize the correlation between the two feature vectors, which can be quantified by the metric of mutual information (MI). 
Inspired by the lower bound of MI~\cite{van2018representation}, the objective to minimize the divergence for $F$ is given by the following contrastive form:
\begin{equation}
    \mathcal{L}_{con.}=-\frac{1}{|\mathcal{H}|^2}\frac{1}{|\mathcal{B}|}\sum_{\substack{1\leq j,k\leq|\mathcal{H}|,\\ j\neq k}}\sum_{i=1}^{|\mathcal{B}|}\log\frac{\exp{\{F_{-1}^\top(\tilde{x}_{ij})F_{-1}(\tilde{x}_{ik})/\tau}\}}{\sum_{u=1}^{|\mathcal{B}|}\exp{\{F_{-1}^\top(\tilde{x}_{ij})F_{-1}(\tilde{x}_{uk})/\tau\}}},\label{eq:con}
\end{equation}
where $\tau$ is a scalar temperature coefficient. 
For the classification problem, we can alternatively adopt the supervised form of the contrastive loss $\mathcal{L}_{con.}$, where $\tilde{x}_{uk}$ with the same class label as $\tilde{x}_{ij}$ are also taken into consideration as positive samples in Eq. \ref{eq:con}. 
The supervised contrastive loss can benefit to increase the correlation of samples from the same class~\cite{khosla2020supervised} for a more reasonable feature representation. 

In addition, the feature space is expected to reflect the task-specific property for a meaningful representation. 
Thus, we also incorporate the task loss $\mathcal{L}_{task}$, \textit{e.g.}, cross-entropy loss in classification tasks, over the synthetic dataset as a supervision signal for $F$. 
In this way, the overall training objective for $F$ is defined as:
\begin{equation}
    \min_{\psi}\mathcal{L}_{F}=\lambda_{con.}\mathcal{L}_{con.}+\lambda_{task}\mathcal{L}_{task},\label{eq:F}
\end{equation}
where $\lambda_{con.}$ and $\lambda_{task}$ are hyper-parameters controlling the weight for each term. 

$F$ acts as an adversary to minimize the divergence between $\tilde{x}_{ij}$ and $\tilde{x}_{ik}$, while the synthetic dataset is expected to maximize it to increase data diversity. 
To this ends, the similarity between $F_{-1}(\tilde{x}_{ij})$ and $F_{-1}(\tilde{x}_{ik})$ becomes one loss term for hallucinator-basis factorization. 
In this paper, we adopt the cosine-similarity and the objective $\mathcal{L}_{cos.}$ is given by:
\begin{equation}
    \mathcal{L}_{cos.}=\frac{1}{|\mathcal{H}|^2}\frac{1}{|\mathcal{B}|}\sum_{\substack{1\leq j,k\leq|\mathcal{H}|,\\ j\neq k}}\sum_{i=1}^{|\mathcal{B}|}\frac{F_{-1}^\top(\tilde{x}_{ij})F_{-1}(\tilde{x}_{ik})}{\Vert F_{-1}(\tilde{x}_{ij})\Vert_2\Vert F_{-1}(\tilde{x}_{ik})\Vert_2}.\label{eq:cos}
\end{equation}
During training, the feature extractor and the factorized components are updated alternately to play this min-max game. 

\subsection{Factorization Training Pipeline}\label{subsec:3}

Following previous paradigms~\cite{zhao2020dataset,zhao2021distribution,cazenavette2022dataset,wang2022cafe}, the synthetic dataset $\mathcal{S}$ is updated in an iterative algorithm. 
In each iteration, we randomly sample a batch of hallucinators and bases and conduct pair-wise combinations. 
The composed images are evaluated by the objective of dataset distillation $\mathcal{L}_{DD}$ and the similarity metric in Eq. \ref{eq:cos}:
\begin{equation}
    \min_{\mathcal{S}}\mathcal{L}_{\mathcal{S}}=\lambda_{DD}\mathcal{L}_{DD}+\lambda_{cos.}\mathcal{L}_{cos.},\label{eq:S}
\end{equation}
where hyper-parameters $\lambda_{DD}$ and $\lambda_{cos.}$ balance the loss. 

Notably, the hallucinator-basis factorization is compatible with a variety of configurations of $\mathcal{L}_{DD}$ by previous arts, which makes it a versatile and effective strategy for DD. 
In this paper, we adopt the trajectories matching loss in  Cazenavette~\etal~\cite{cazenavette2022dataset} as $\mathcal{L}_{DD}$ by default thanks to its superior performance. 
The basic idea is to update a downstream model from a cached checkpoint $\phi^*_t$ at iteration $t$, using the synthetic dataset $\mathcal{S}$ for $N$ times, and using the real dataset $\mathcal{T}$ for $M$ times respectively. 
The updated parameters by the two cases, $\hat{\phi}_{t+N}$ and $\phi^*_{t+M}$ are enforced to be consistent:
\begin{equation}
    \begin{aligned}
    \hat{\phi}_{t+n+1}\leftarrow\hat{\phi}_{t+n}-\alpha&\nabla_{\hat{\phi}_{t+n}}\mathcal{L}_{task}(\mathcal{S}), \hspace{3mm} \hat{\phi}_t\leftarrow\phi^*_t, \hspace{3mm} 0\leq n<N, \\
    \phi^*_{t+m+1}\leftarrow\phi^*_{t+m}&-\beta\nabla_{\phi^*_{t+m}}\mathcal{L}_{task}(\mathcal{T}), \hspace{3mm} 0\leq m<M, \\
    \mathcal{L}_{DD}&=\frac{\Vert\hat{\phi}_{t+N}-\phi^*_{t+M} \Vert_2^2}{\Vert\phi^*_t-\phi^*_{t+M}\Vert_2^2},&
    \end{aligned}
\end{equation}
where $\alpha$ and $\beta$ are learning rates with $\mathcal{S}$ and $\mathcal{T}$ respectively. 
$\alpha$ is learnable in the framework while $\beta$ is a hyper-parameter. 
In Sec. \ref{sec:4-2}, we also experiment with other settings of $\mathcal{L}_{DD}$. 

Based on the supervised signals in Eq. \ref{eq:S}, the gradients are backward propagated to the composed images and finally to the sampled hallucinators and bases so as to be updated using a decent algorithm such as SGD. 
Since all the operations are differentiable, the training can be completed end-to-end. 

\section{Experiments}\label{sec:4}

\subsection{Datasets and Implementing Details}\label{sec:4-1}

We conduct evaluations of our method on three standard image classification benchmarks: SVHN~\cite{netzer2011reading}, CIFAR10, and CIFAR100~\cite{krizhevsky2009learning}. 
There are 60,000 images for real-world digit recognition in SVHN. 
For CIFAR10 and CIFAR100, there are 50,000 training images in total.
The number of classes for the three datasets are $10$, $10$, and $100$ respectively. 
All the images are under $32\times32$ resolution in 3-channel RGB format. 
Following previous works~\cite{cazenavette2022dataset}, we use ZCA for image preprocessing with Kornia implementation~\cite{riba2020kornia} before all the experiments. 
Experiments with more datasets, including images in larger spatial scales, can be found in the supplement. 

In this paper, for convenience of comparisons with prior works, we maintain the same size with images in original datasets, \textit{i.e.}, $h'=h$, $w'=w$, and $c'=3$ for bases. 
We also experiment with other sizes of bases in Sec. \ref{sec:4-3}. 
For hallucinators, the encoder and decoder contain $1$ \texttt{Conv-ReLU} blocks. 
The number of feature channel $c''$ is $3$. 
We use 5 hallucinators by default. 
The learning rates of hallucinators and bases, $\eta_H$ and $\eta_B$, are the same and for the feature extractor, the learning rate $\eta_F$ is $0.001$. 
Hyper-parameters $\lambda_{con.}$, $\lambda_{task}$, $\lambda_{DD}$, and $\lambda_{cos.}$ are set as $0.1$, $1$, $1$, and $0.1$ empirically. 
Sensitivities of these hyper-parameters are analyzed in Sec. \ref{sec:4-3}. 
The adversary network has the same architecture as that for computing $\mathcal{L}_{DD}$. 
In experiments on SVHN and CIFAR10, we incorporate all the bases in each iteration, while in experiments on CIFAR100, we adopt a batch size of 300 when the total number of bases is greater than 1,000. 
We only consider random 2 hallucinators in one iteration for simplicity. 
The maximal configuration of computational resources is 4 24GB 3090 GPUs. 
The GPU memory consumption is dependent on that of the baseline method for $\mathcal{L}_{DD}$ and is slightly higher than it due to the computation of $\mathcal{L}_{cos.}$ and $\mathcal{L}_{con.}$. 
The baseline method for $\mathcal{L}_{DD}$ is MTT~\cite{cazenavette2022dataset} if not specified. 
Other settings related to DD hold the same as the baseline. 
All the quantitative results are based on the mean and standard deviation over 5 repeated experiments. 
\xw{To make sure fair comparisons, the dataset size in our method is equal to the number of bases $|\mathcal{B}|$ and the hallucinators are treated as parameterized data augmentors working online in downstream training, just as general data augmentations, which means that the dataset size does not increase compared with the baselines.}

\subsection{Comparisons}

\begin{table}[!t]
\begin{floatrow}[1]
\tableboxf{}{
    \centering
    \scriptsize
    \setlength{\tabcolsep}{2pt}
    \vspace{-5pt}
    \begin{tabular}{cc|ccc|ccc|ccc}
    \toprule
    \multirow{3}{*}{} & Dataset & \multicolumn{3}{c|}{SVHN} & \multicolumn{3}{c|}{CIFAR10} & \multicolumn{3}{c}{CIFAR100} \\
    \midrule
    & IPC & 1 & 10 & 50 & 1 & 10 & 50 & 1 & 10 & 50 \\
    & Ratio \% & 0.014 & 0.14 & 0.7 & 0.02 & 0.2 & 1 & 0.2 & 2 & 10 \\ \midrule
    \multirow{4}{*}{Coreset} & Random & 14.6$\pm$1.6 & 35.1$\pm$4.1 & 70.9$\pm$0.9 & 14.4$\pm$2.0 & 26.0$\pm$1.2 & 43.4$\pm$1.0 & 4.2$\pm$0.3 & 14.6$\pm$0.5 & 30.0$\pm$0.4  \\
    & Herding & 20.9$\pm$1.3 & 50.5$\pm$3.3 & 72.6$\pm$0.8 & 21.5$\pm$1.3 & 31.6$\pm$0.7 & 40.4$\pm$0.6 & 8.4$\pm$0.3 & 17.3$\pm$0.3 & 33.7$\pm$0.5  \\
    & K-Center & 21.0$\pm$1.5 & 14.0$\pm$1.3 & 20.1$\pm$1.4 & 21.5$\pm$1.3 & 14.7$\pm$0.9 & 27.0$\pm$1.4 & 8.3$\pm$0.3 & 7.1$\pm$0.2 & 30.5$\pm$0.3 \\
    & Forgetting & 12.1$\pm$1.7 & 16.8$\pm$1.2 & 27.2$\pm$1.5 & 13.5$\pm$1.2 & 23.3$\pm$1.0 & 23.3$\pm$1.1 & 4.5$\pm$0.3 & 9.8$\pm$0.2 & -  \\
    \midrule
    \multirow{8}{*}{Distillation} & DD$^{\dag}$~\cite{wang2018dataset} & - & - & - & - & 36.8$\pm$1.2 & - & - & - & -  \\
    & LD$^{\dag}$~\cite{bohdal2020flexible} & - & - & - & 25.7$\pm$0.7 & 38.3$\pm$0.4 & 42.5$\pm$0.4 & 11.5$\pm$0.4 & - & -  \\
    & DC~\cite{zhao2020dataset} & 31.2$\pm$1.4 & 76.1$\pm$0.6 & 82.3$\pm$0.3 & 28.3$\pm$0.5 & 44.9$\pm$0.5 & 53.9$\pm$0.5 & 12.8$\pm$0.3 & 25.2$\pm$0.3 & -  \\
    & DSA~\cite{zhao2021dataset} & 27.5$\pm$1.4 & 79.2$\pm$0.5 & 84.4$\pm$0.4 & 28.8$\pm$0.7 & 52.1$\pm$0.5 & 60.6$\pm$0.5 & 13.9$\pm$0.3 & 32.3$\pm$0.3 & 42.8$\pm$0.4  \\
    & DM~\cite{zhao2021distribution} & - & - & - & 26.0$\pm$0.8 & 48.9$\pm$0.6 & 63.0$\pm$0.4 & 11.4$\pm$0.3 & 29.7$\pm$0.3 & 43.6$\pm$0.4  \\
    & CAFE~\cite{wang2022cafe} & 42.6$\pm$3.3 & 75.9$\pm$0.6 & 81.3$\pm$0.3 & 30.3$\pm$1.1 & 46.3$\pm$0.6 & 55.5$\pm$0.6 & 12.9$\pm$0.3 & 27.8$\pm$0.3 & 37.9$\pm$0.3  \\
    & CAFE+DSA~\cite{wang2022cafe} & 42.9$\pm$3.0 & 77.9$\pm$0.6 & 82.3$\pm$0.4 & 31.6$\pm$0.8 & 50.9$\pm$0.5 & 62.3$\pm$0.4 & 14.0$\pm$0.3 & 31.5$\pm$0.2 & 42.9$\pm$0.2  \\
    & MTT~\cite{cazenavette2022dataset} & \underline{58.5$\pm$1.4} & \underline{70.8$\pm$1.8} & \underline{85.7$\pm$0.1} & 46.3$\pm$0.8 & 65.3$\pm$0.7 & 71.6$\pm$0.2 & 24.3$\pm$0.3 & \underline{39.0$\pm$0.1} & \underline{46.1$\pm$0.2}  \\
    \midrule
    \multirow{3}{*}{Factorization} & BPC & 1 & 9 & 49 & 1 & 9 & 49 & 1 & 9 & 49 \\
    & Ratio \% & 0.028 & 0.14 & 0.7 & 0.04 & 0.2 & 1 & 0.22 & 1.82 & 9.82 \\
    & HaBa & \textbf{69.8$\pm$1.3} & \textbf{83.2$\pm$0.4} & \textbf{88.3$\pm$0.1} & \textbf{48.3$\pm$0.8} & \textbf{69.9$\pm$0.4} & \textbf{74.0$\pm$0.2} & \textbf{33.4$\pm$0.4} & \textbf{40.2$\pm$0.2} & \textbf{47.0$\pm$0.2}  \\
    \midrule
    \multicolumn{2}{c|}{Whole Dataset} & \multicolumn{3}{c|}{95.4$\pm$0.1} & \multicolumn{3}{c|}{84.8$\pm$0.1} & \multicolumn{3}{c}{56.2$\pm$0.3}   \\
    \bottomrule
    \end{tabular}
    \vspace{-0.3cm}
    \caption{The performance (test accuracy \%) comparison to state-of-the-art methods. LD$^{\dag}$ and DD$^{\dag}$ use AlexNet for CIFAR10, while the rest use ConvNet for training and testing. IPC: Number of Images Per Class; BPC: Number of Bases Per Class; Ratio~(\%): the ratio of distilled images to whole training set. Underline denotes results by our implementation.}
\label{tab:compare_sota}
}
\end{floatrow}
\vspace{-0.5cm}
\end{table}

\textbf{Comparisons with State of the Arts:} \label{sec:4-2}
We compare HaBa with previous state of the arts for DD in standard settings, to synthesize 1, 10, and 50 images per class (IPC) respectively. 
In our setting, the number of parameters in a hallucinator is is approximately equal to that for 2 synthetic images, while the size of a basis is equal to that of an image. 
Taking the storage cost of 5 hallucinators into consideration, we set the number of bases per class (BPC) as IPC minus 1 in each IPC configuration when IPC is greater than 1, to make the comparisons as fair as possible. 
Candidates are coreset based methods including Random~\cite{chen2012super,rebuffi2017icarl}, Herding~\cite{castro2018end,belouadah2020scail}, K-Center~\cite{farahani2009facility,sener2017active}, and Forgetting~\cite{toneva2018empirical}, meta learning based methods including DD ~\cite{wang2018dataset} and LD~\cite{bohdal2020flexible}, training matching based methods including DC~\cite{zhao2020dataset}, DSA~\cite{zhao2021dataset}, and MTT~\cite{cazenavette2022dataset}, and distribution matching based methods including DM~\cite{zhao2021distribution} and CAFE~\cite{wang2022cafe}. 
The comparisons follow the standard protocol adopting a 3-layer \texttt{Conv-InstanceNorm-ReLU-AvgPool} ConvNet with 128 channels in training and testing. 

The comparison results are shown in Tab. \ref{tab:compare_sota} and we can observe that HaBa achieves state-of-the-art performance in all datasets and settings. 
Especially when the ratio of distilled images to the whole training set is less than 1\%, our method can yield significant improvement over all the candidate methods, which demonstrates that the scheme of hallucinator-basis factorization improves the data efficiency for the task of dataset distillation. 

\begin{figure}[!t]
\begin{floatrow}[1]
\figureboxf{}{
  \centering
  \begin{subfigure}{0.245\linewidth}
  \includegraphics[width=\linewidth]{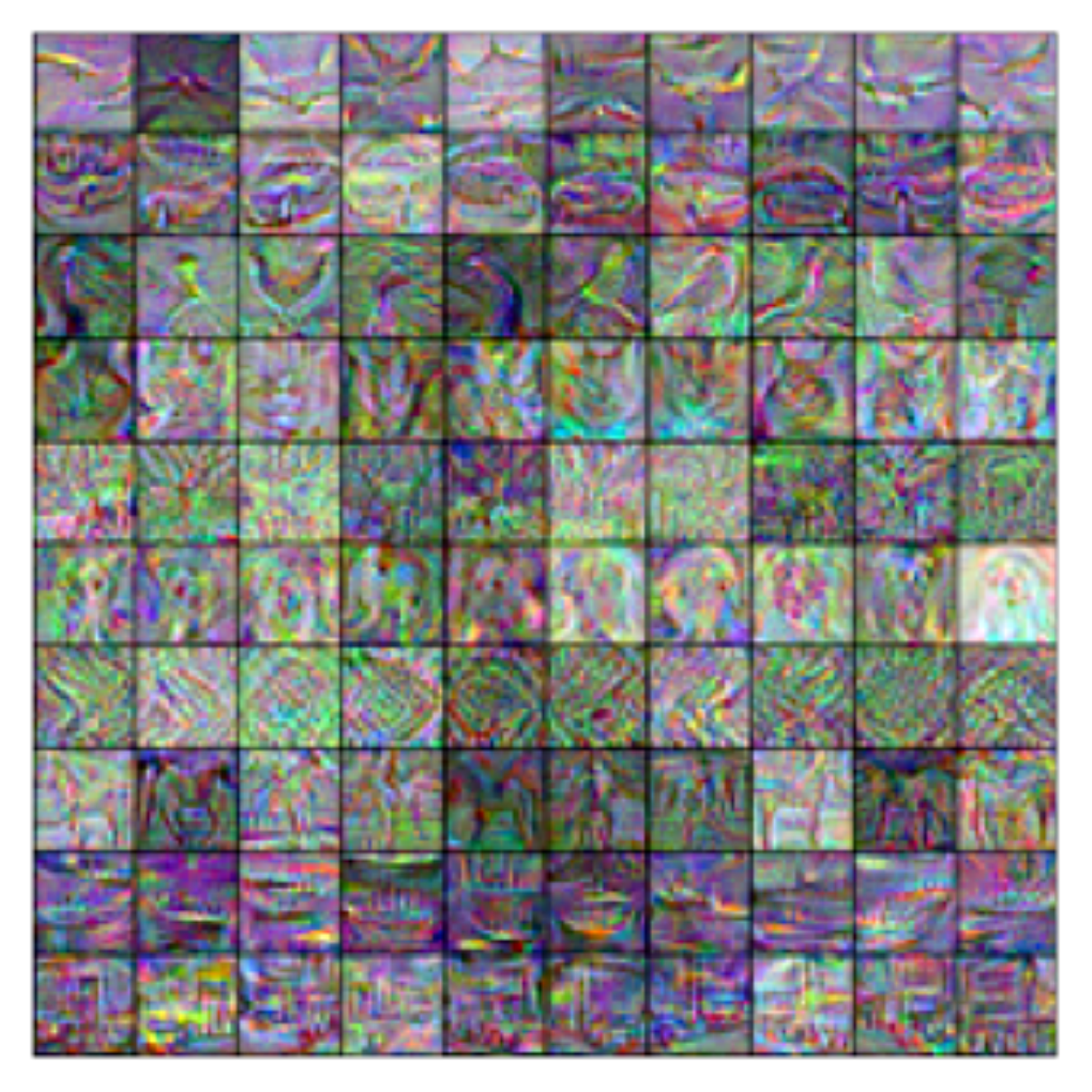}
  \vspace{-0.7cm}
  \caption{Bases}
  \end{subfigure}
  \begin{subfigure}{0.245\linewidth}
  \includegraphics[width=\linewidth]{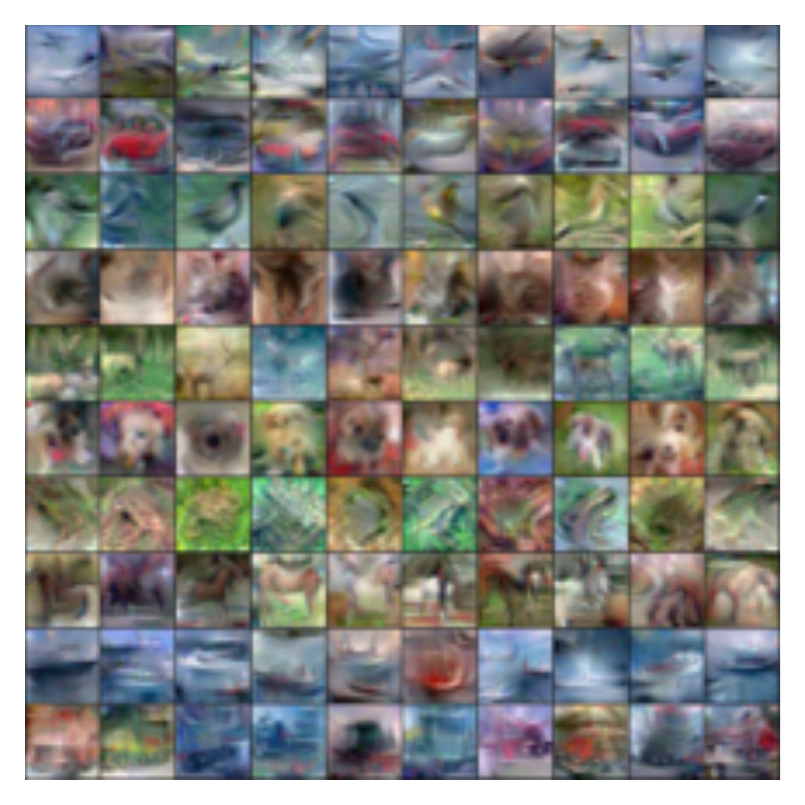}
  \vspace{-0.7cm}
  \caption{Images by $H_1$}
  \end{subfigure}
  \begin{subfigure}{0.245\linewidth}
  \includegraphics[width=\linewidth]{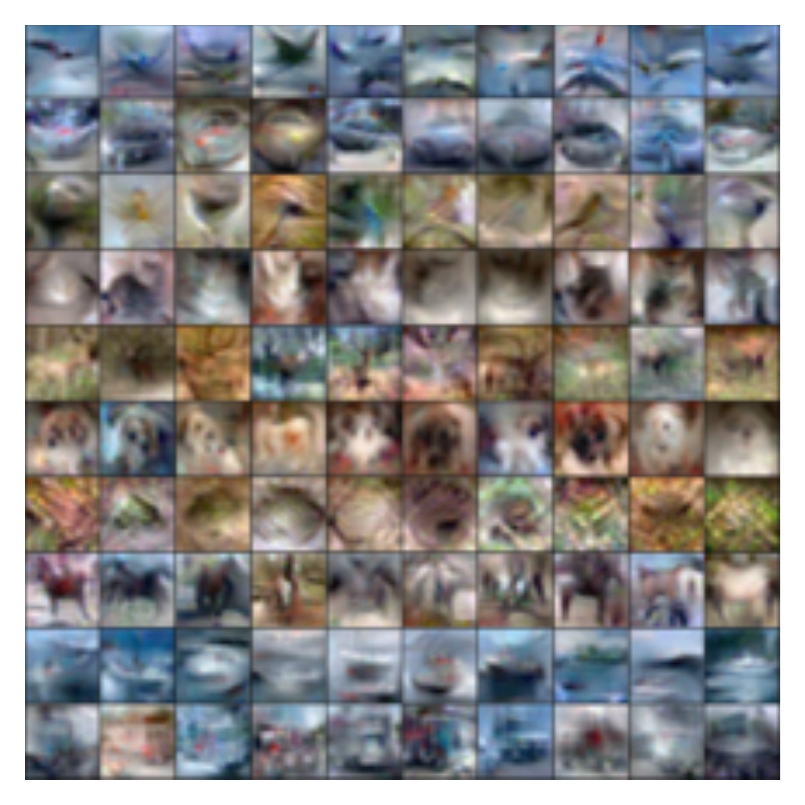}
  \vspace{-0.7cm}
  \caption{Images by $H_2$}
  \end{subfigure}
  \begin{subfigure}{0.245\linewidth}
  \includegraphics[width=\linewidth]{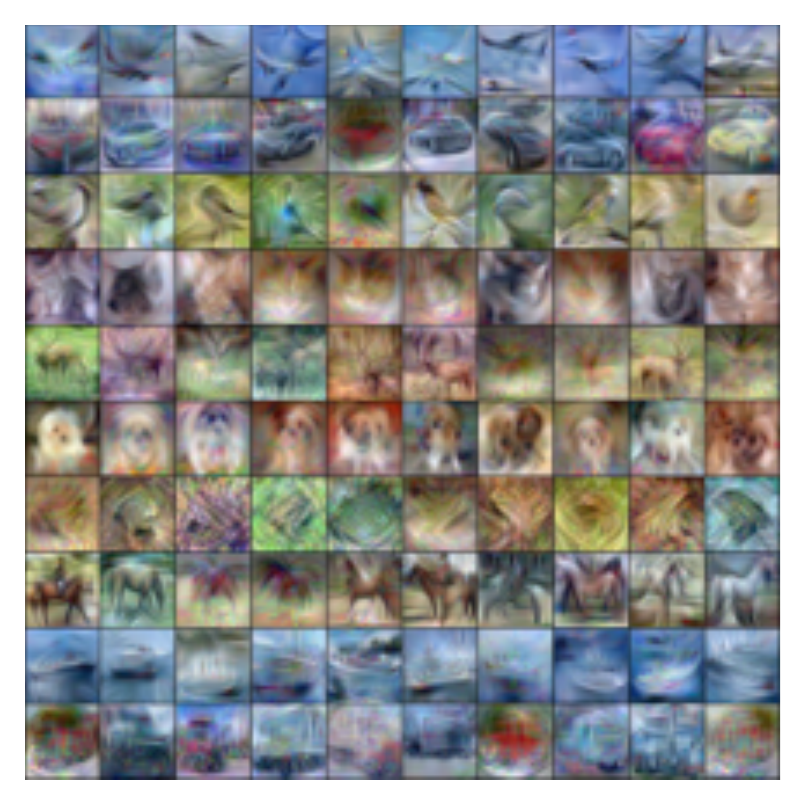}
  \vspace{-0.7cm}
  \caption{Images by Baseline}
  \end{subfigure}
  \vspace{-0.3cm}
  \caption{Visualization of factorized results by our HaBa (70.27\% test acc.) and baseline MTT (65.92\% test acc.). Zoom-in for better comparisons.}
  \label{fig:visualization}
  }
   \vspace{-0.8cm}
  \end{floatrow}
\end{figure}

\textbf{Qualitative Comparisons:} 
We visualize the factorized results by our method as well as the baseline on CIFAR10 dataset with 10 BPC in Fig. \ref{fig:visualization}. 
Due to the space limitation, we only provide images generated by 2 hallucinators here. 
More results can be found in the supplement. 
As shown in the figure, we can find that bases mainly store some main structures and contour information. 
Different hallucinators would render a basis with diverse styles and details. 
Thanks to the dataset factorization scheme, the diversity of distilled images by our method is higher than that by the baseline. 

\textbf{Building upon Different Baselines:} 
To reflect the versatility of the insight, we implement HaBa on multiple state-of-the-art training pipelines of DD, including DC, DM, and MTT. 
We evaluate the performance of synthetic datasets on CIFAR10 and maintain the IPC of baseline methods as BPC plus 1, which makes storage costs for synthetic datasets as close as possible for fairness. 
As shown in Tab. \ref{tab:cross_architecture}, when training and testing on ConvNet, the strategy of HaBa can make a consistent improvement over all the baselines, which demonstrates that factorization is a general idea to improve the data efficiency in DD. 

\begin{table}[!t]
\begin{floatrow}[1]
\tableboxf{}{
    \centering
    \scriptsize
    \setlength{\tabcolsep}{1pt}
    \vspace{-5pt}
    \begin{tabular}{c|c|ccc|ccc|ccc}
    \toprule
    \multirow{3}{*}{} & Method & \multicolumn{3}{c|}{DC~\cite{zhao2020dataset}} & \multicolumn{3}{c|}{DM~\cite{zhao2021distribution}} & \multicolumn{3}{c}{MTT~\cite{cazenavette2022dataset}} \\
    \midrule
    & IPC & 2 & 11 & 51 & 2 & 11 & 51 & 2 & 11 & 51 \\
    & BPC & 1 & 10 & 50 & 1 & 10 & 50 & 1 & 10 & 50 \\ 
    \midrule
    \multirow{3}{*}{ConvNet} & Baseline & 31.36$\pm$0.16 & 45.29$\pm$0.30 & 54.24$\pm$0.61 & 34.57$\pm$0.52 & 50.35$\pm$0.36 & 62.03$\pm$0.29 & 50.59$\pm$0.95 & 63.90$\pm$0.29 & 69.81$\pm$0.48 \\
    & w. HaBa & 34.11$\pm$0.47 & 49.88$\pm$0.52 & 58.91$\pm$0.23 & 37.32$\pm$0.13 & 56.83$\pm$0.11 & 64.44$\pm$0.40 & 56.76$\pm$0.38 & 69.48$\pm$0.26 & 73.25$\pm$0.21 \\
    & Gain & +2.75 & +4.59 & +4.67 & +2.75 & +6.48 & +2.41 & +6.17 & +5.58 & +3.44 \\
    \midrule
    \multirow{3}{*}{ResNet} & Baseline & 18.10$\pm$0.76 & 18.36$\pm$0.36 & 22.14$\pm$0.38 & 22.25$\pm$1.00 & 40.00$\pm$1.49 & 53.40$\pm$0.68 & 35.15$\pm$0.96 & 45.05$\pm$1.46 & 54.47$\pm$0.95 \\
    & w. HaBa & 24.49$\pm$0.55 & 24.27$\pm$0.56 & 31.08$\pm$0.32 & 31.34$\pm$0.72 & 47.57$\pm$0.49 & 59.61$\pm$0.35 & 47.39$\pm$0.71 & 57.97$\pm$0.88 & 64.35$\pm$0.60 \\
    & Gain & +6.39 & +6.11 & +8.94 & +9.09 & +7.57 & +6.21 & +12.24 & +12.92 & +9.88 \\
    \midrule
    \multirow{3}{*}{VGG} & Baseline & 28.02$\pm$0.26 & 35.88$\pm$0.67 & 38.73$\pm$0.48 & 22.28$\pm$1.03 & 41.64$\pm$0.64 & 55.17$\pm$0.54 & 38.04$\pm$1.19 & 50.49$\pm$1.02 & 61.36$\pm$0.30 \\
    & w. HaBa & 29.42$\pm$0.93 & 37.03$\pm$0.42 & 41.91$\pm$0.55 & 26.93$\pm$0.62 & 49.41$\pm$0.36 & 67.47$\pm$0.43 & 48.26$\pm$0.54 & 60.47$\pm$0.56 & 67.47$\pm$0.43 \\
    & Gain & +1.40 & +1.15 & +3.18 & +4.65 & +7.77 & +12.30 & +10.22 & +9.98 & +6.11 \\
    \midrule
    \multirow{3}{*}{AlexNet} & Baseline & 20.02$\pm$1.31 & 22.42$\pm$1.35 & 29.48$\pm$0.87 & 20.67$\pm$3.64 & 37.04$\pm$0.92 & 49.14$\pm$0.94 & 26,06$\pm$1.01 & 35.95$\pm$1.52 & 49.20$\pm$1.27 \\
    & w. HaBa & 22.24$\pm$1.14 & 33.02$\pm$0.91 & 33.42$\pm$1.39 & 32.14$\pm$0.60 & 44.14$\pm$0.67 & 53.09$\pm$0.89 & 43.63$\pm$1.46 & 48.96$\pm$3.00 & 60.07$\pm$1.37 \\
    & Gain & +2.22 & +10.60 & +3.94 & +11.47 & +7.10 & +3.95 & +17.57 & +13.01 & +10.87 \\
    \bottomrule
    \end{tabular}
    \vspace{-0.2cm}
    \caption{Cross-architecture performance (test accuracy \%) comparison to different baseline methods of DD HaBa built upon.}
\label{tab:cross_architecture}
}
\end{floatrow}
\vspace{-0.8cm}
\end{table}

\textbf{Cross-Architecture Performance:} 
For DD, a satisfactory distilled dataset should have similar training effects to the original one on downstream models with arbitrary architectures. 
Thus, cross-architecture generalization performance is an important metric for DD. 
We use the synthetic datasets trained on ConvNet to train models with different structures including ResNet~\cite{he2016deep}, VGG~\cite{simonyan2014very}, and AlexNet~\cite{krizhevsky2012imagenet}. 
The results can be found in Tab. \ref{tab:cross_architecture}. 
\xw{Benefiting from the increased data diversity}, HaBa can improve the across-architecture accuracy significantly with a performance gain up to $17.57\%$. 
The consistent and significant improvement validates the superior ability of our method to \xw{capture the informative features} and thus original datasets can be replaced by the synthetic ones better. 

\textbf{Comparisons under the Same Number of Final Images:}
\xw{
In the default comparison protocol, we compare our method with the baselines using the same budget of storage, where our method can store information of exponentially more images than the baselines with the same number of parameters. 
In this part, we also examine the performance of HaBa under the condition that the number of final images, \textit{i.e.}, $|\mathcal{H}|\times|\mathcal{B}|$, is equal to that used by the baseline. 
Intuitively, given that the objective functions of our method and the baseline are the same exactly, the performance of the baseline can be viewed as an upper bound of ours, since there are significantly less parameters in our method to carry the information of final images in this case. 
Therefore, we first remove the term $\mathcal{L}_{cos.}$ from the loss function of DD in Eq. \ref{eq:S} to guarantee a consistent optimization objective with the baseline. 
Then, we compare the performance of HaBa and the baseline using 10, 20, 30, 40, and 50 final images respectively. 
Here, the number of hallucinators $|\mathcal{H}|$ is 2 and the number of bases is thus half of the number of final images. 
As shown in the red and green curves in Fig. \ref{fig:same_img}, performance of the baseline can be well approximated by ours with only half of the number of parameters, especially when the number of images is relatively large. 
Remarkably, with the proposed adversary contrastive constraint, our method can even outperform the baseline consistently, as shown in the blue curve, which further demonstrates the effectiveness of the proposed solution. 
}

\begin{figure}[!t]
\begin{floatrow}[2]

\tableboxl{\caption{\xw{Comparisons with concurrent work IDC~\cite{kim2022dataset} on efficient synthetic parameterization.}}}{
    \centering
    \scriptsize
    \setlength{\tabcolsep}{2pt}
    \vspace{-5pt}
    \begin{tabular}{cccc}
        \toprule
        \# of Param. / Class & 2$\times$32$\times$32$\times$3 & 11$\times$32$\times$32$\times$3 & 51$\times$32$\times$32$\times$3 \\ 
        \midrule
        Baseline & 49.89$\pm$0.95 & 65.92$\pm$0.62 & 70.73$\pm$0.52 \\
        w. IDC & 56.13$\pm$0.38 & 70.85$\pm$0.43 & 71.01$\pm$0.41 \\
        w. IDC \& HaBa & 61.27$\pm$0.34 & 72.14$\pm$0.22 & 75.31$\pm$0.27 \\
        \bottomrule
    \end{tabular}
    \label{tab:idc}
}

\figureboxl{\caption{Comparisons on the setting of continual setting. Results on the ConvNet3 (Left) and ResNet18 (Right) architectures are shown.}}{
    \includegraphics[width=0.99\linewidth]{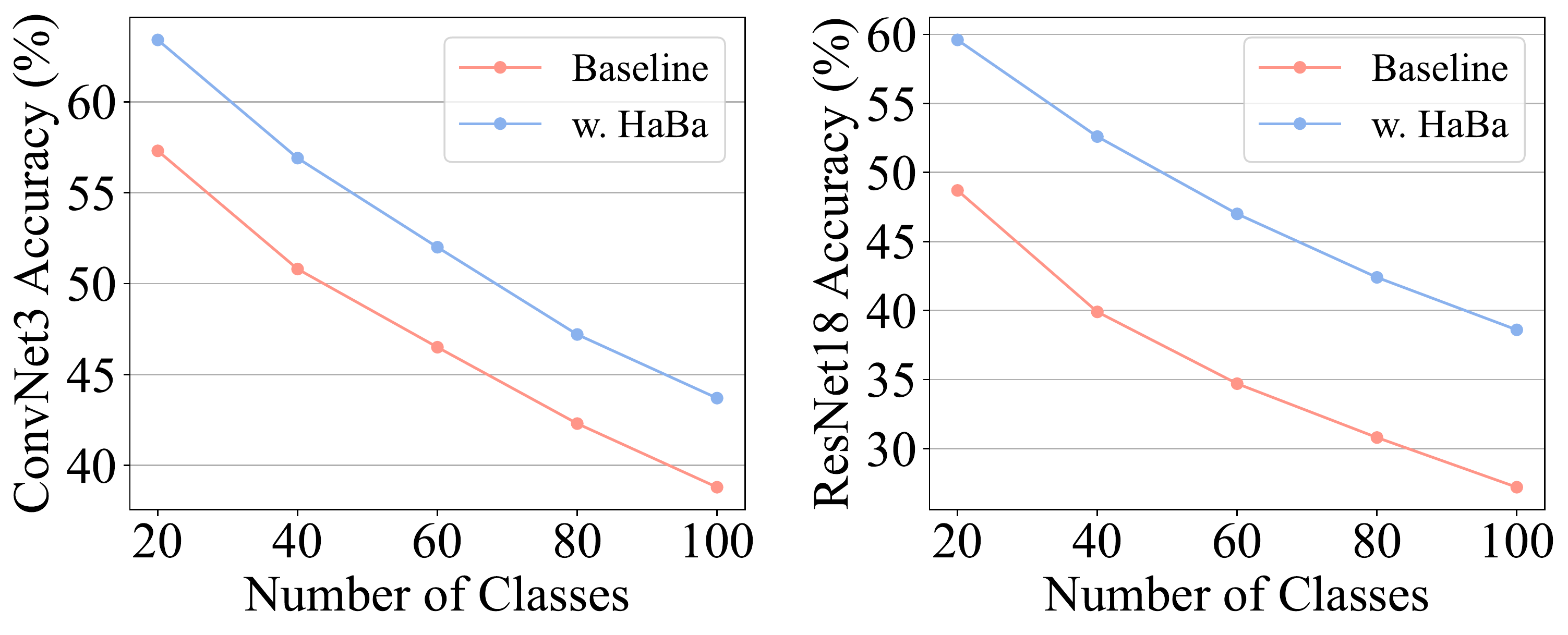}
    \vspace{0.1cm}
    \label{fig:cl}
    \vspace{-0.8cm}
}

\end{floatrow}
\vspace{-0.5cm}
\end{figure}

\textbf{Comparisons with Concurrent Works on Efficient Distilled Dataset Parameterization:} 
As a concurrent work on efficient distilled dataset parameterization, IDC~\cite{kim2022dataset} is proposed to store $4\times$ number of $2\times$ down-sample synthetic images compared with the baseline. 
The core is to reduce the spatial size for efficient parameterization. 
For HaBa of this paper, instead, we do not modify the spatial size of bases in the default setting for better qualitative explainablity and more intuitive comparisons with the baselines. 
In this sense, IDC and HaBa are in fact two orthogonal techniques and they can readily join force to enhance the baseline performance. 
Here, we try using the technique of IDC and adopting $2\times$ down-sample synthetic images on the baseline MTT, based on which we further consider adding our HaBa and involving 5 hallucinators. 
As shown in Tab. \ref{tab:idc}, with the efficient parameterization of IDC, the performance of baseline can be improved. 
With HaBa in this paper, the performance can even be further improved a lot: 5.14\%, 1.29\%, and 4.30\% in the three settings respectively, which demonstrates that IDC and HaBa work in different ways. 

\textbf{Applications in Continual Learning:} 
To further demonstrate the advantage of the proposed method for improving data efficiency, following the setting of DM~\cite{zhao2021distribution}, we conduct experiments on the setting of continual learning on CIFAR-100, with 20 random classes per stage. 
The average number of parameters per class is $20\times32\times32\times3$. 
The synthetic datasets are trained with a ConvNet with 3 blocks. 
We evaluate synthetic datasets by our method and the DM baseline on the same ConvNet architecture and ResNet18. 
The results in Fig. \ref{fig:cl} demonstrate that the proposed method increases the informativeness of synthetic datasets and thus produce significantly better performance, especially in the cross-architecture setting. 

\subsection{Ablation Studies}\label{sec:4-3}

\begin{figure}[!t]
\begin{floatrow}[2]
\tablebox{\caption{Results of ablation study on loss terms in HaBa: $\mathcal{L}_{cos.}$, $\mathcal{L}_{con.}$, and $\mathcal{L}_{task}$.}}{
    \centering
    \scriptsize
    \setlength{\tabcolsep}{2pt}
    \vspace{-5pt}
    \begin{tabular}{cccc}
        \toprule
        BPC & 1 & 10 & 50 \\ 
        \midrule
        HaBa w/o $\mathcal{L}_{cos.}$ & 54.56$\pm$0.61 & 70.16$\pm$0.44 & 73.93$\pm$0.21 \\
        HaBa w/o $\mathcal{L}_{con.}$ & 54.91$\pm$0.49 & 70.07$\pm$0.48 & 72.50$\pm$0.39 \\
        HaBa w/o $\mathcal{L}_{task}$ & 54.62$\pm$0.42 & 70.07$\pm$0.16 & 72.74$\pm$0.20 \\
        \midrule
        HaBa Full & 55.66$\pm$0.29 & 70.27$\pm$0.63 & 74.04$\pm$0.16 \\
        \midrule
        HaBa w $\mathcal{L}_{con.}$ Downstream & 56.78$\pm$0.22 & 70.44$\pm$0.15 & 75.00$\pm$0.52 \\
        \bottomrule
    \end{tabular}
    \label{tab:ablation}
}
\figurebox{\caption{Impacts of different $\lambda_{cos.}$ and $\lambda_{con.}$ on the test accuracy.}}{
    \includegraphics[width=0.99\linewidth]{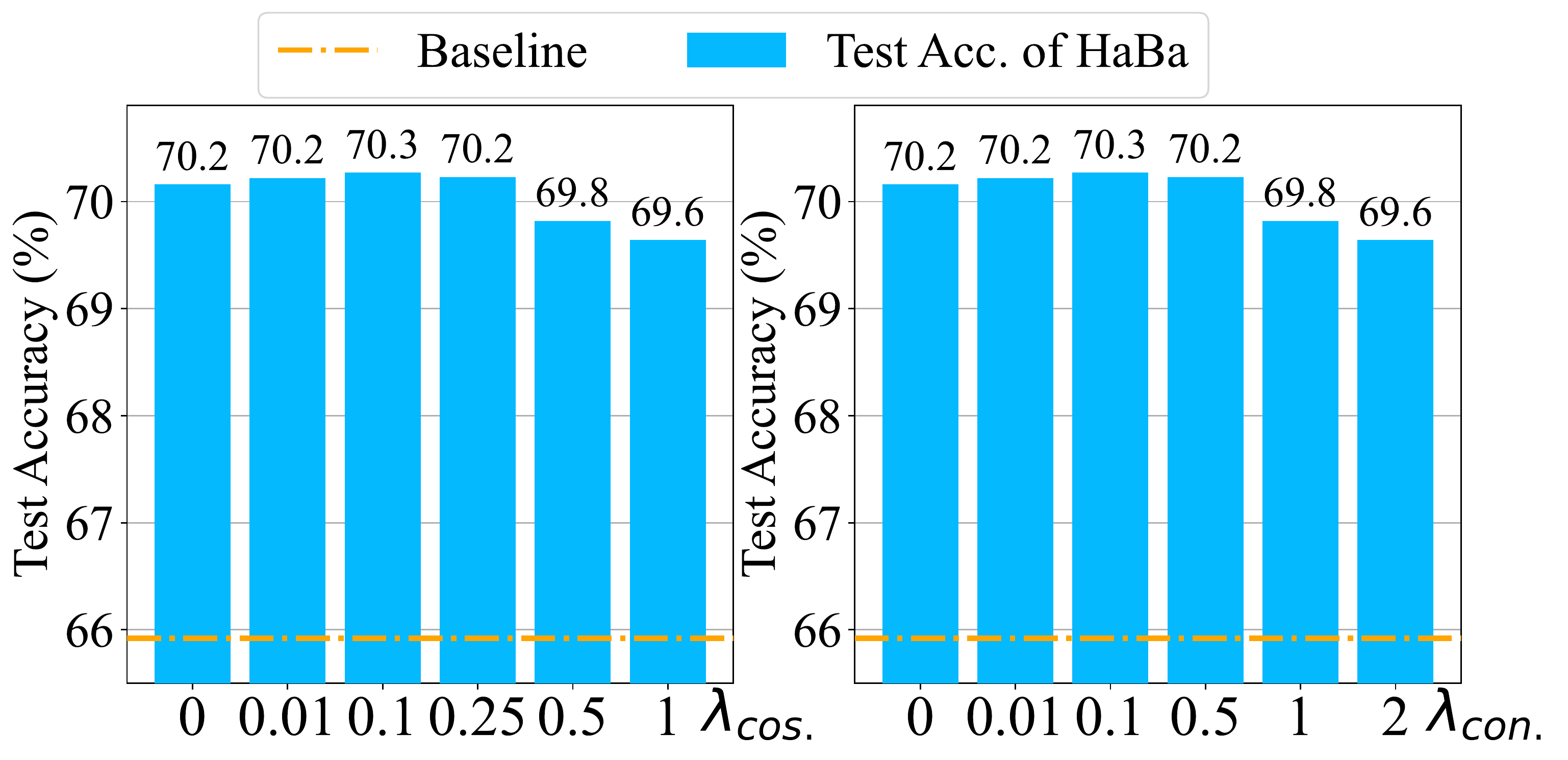}
    \label{fig:sensitivity}
    \vspace{-0.6cm}
}
\end{floatrow}
\vspace{-0.8cm}
\end{figure}

\textbf{Loss Terms:} 
To validate the effectiveness of the proposed adversarial contrastive \xw{constraints}, we design ablation studies on the CIFAR10 dataset over three loss terms: $\mathcal{L}_{cos.}$ in Eq. \ref{eq:cos}, $\mathcal{L}_{con.}$ in Eq. \ref{eq:con}, and the task-specific loss $\mathcal{L}_{task}$. 
Through the results in Tab. \ref{tab:ablation}, we can find that deleting any one of them would hurt the performance. 
We also experiment with involving $\mathcal{L}_{con.}$ for downstream training, to enforce the similarity among images composed of different hallucinators and a common basis. 
Observed from the last row of Tab. \ref{tab:ablation}, the performance can be further improved, since $\mathcal{L}_{con.}$ helps the representation learning of related samples~\cite{khosla2020supervised}. 
Note that we do not use this loss term for downstream training in other experiments for fair and standard comparison. 

We examine the sensitivities of hyper-parameters $\lambda_{cos.}$ and $\lambda_{con.}$ used to balance the weights of loss terms $\mathcal{L}_{cos.}$ and $\mathcal{L}_{con.}$ respectively in Fig. \ref{fig:sensitivity}. 
The results are evaluated on the CIFAR10 dataset with 10 BPC. 
We can observe that the overall performance is not sensitive to the selection of these hyper-parameters and our method makes a consistent improvement over the baseline with 11 IPC. 

\textbf{Class-Independent Hallucinators v.s. Shared Hallucinators:} 
In the default setting of HaBa, each class maintains a certain number of bases independently and all the classes share the same set of hallucinators. 
But what about the case that hallucinators are also made class-independent?
We study this problem experimentally in Tab. \ref{tab:share_hallucinator}. 
Given the same BPC, class-independent hallucinators can indeed somehow improve the performance when there are fewer synthetic samples, \textit{e.g.}, 1 BPC. 
However, when BPC is higher, equipping each class with an independent set of hallucinators would not benefit the performance. 
There are probably two reasons: (1) shared hallucinators across all the classes extract global information of the whole dataset, which encodes more representative and universal knowledge; and (2) the class-independent case would make the number of hallucinators 10 times for the CIFAR10 dataset, which leaves a heavy burden for the optimization process. 
Thus, as indicated in Tab. \ref{tab:share_hallucinator}, a better solution is to make room for more bases using the memory allocated to store class-independent hallucinators initially, which would result in more satisfactory data efficiency. 

\begin{figure}[!t]
\begin{floatrow}[2]

\figurebox{\caption{\xw{Comparisons with the baseline under the same number of final images.}}}{
    \includegraphics[width=0.85\linewidth]{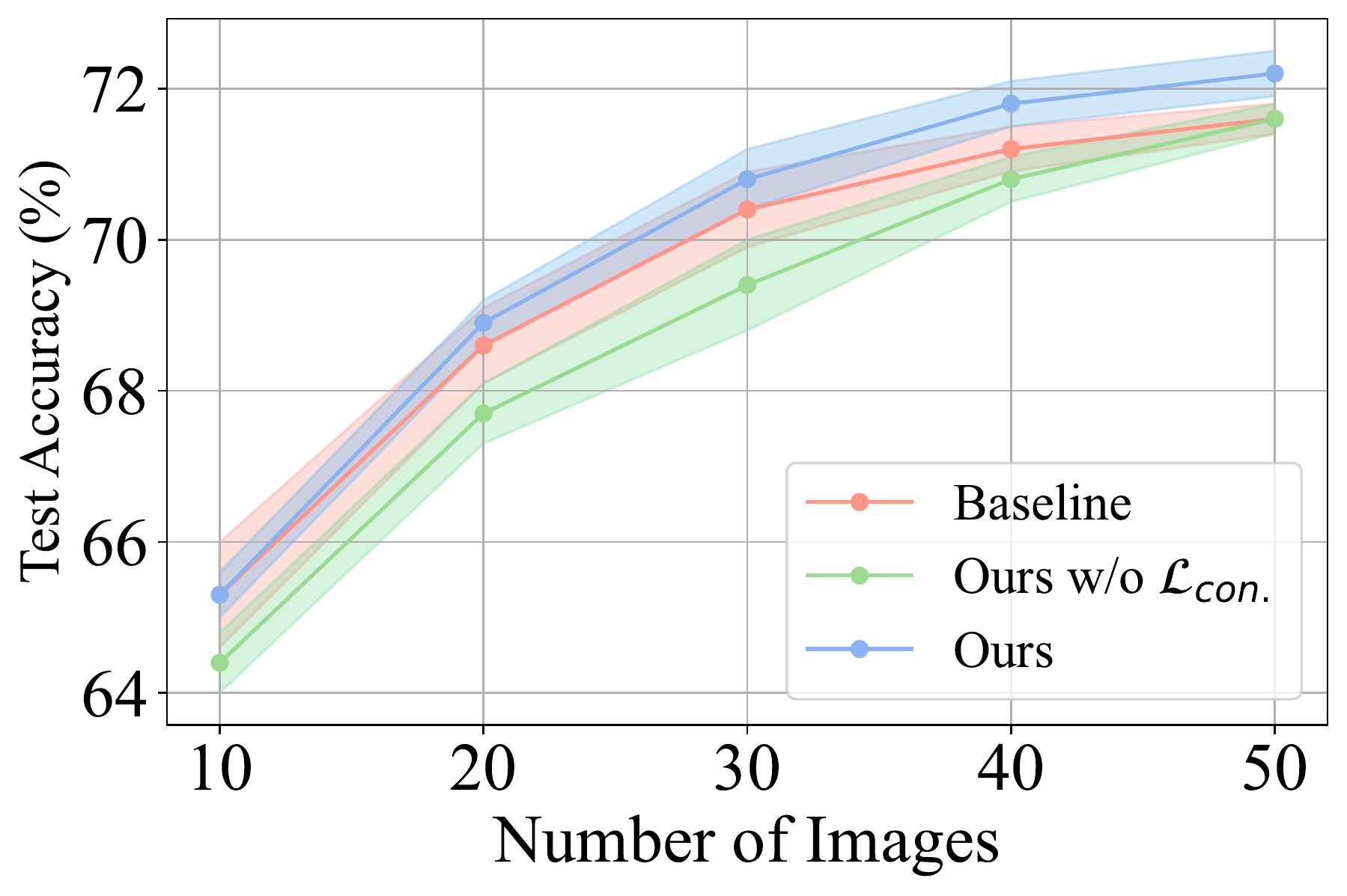}
    \label{fig:same_img}
    \vspace{-0.4cm}
}
\figurebox{}{
    \centering
    \includegraphics[width=0.99\linewidth]{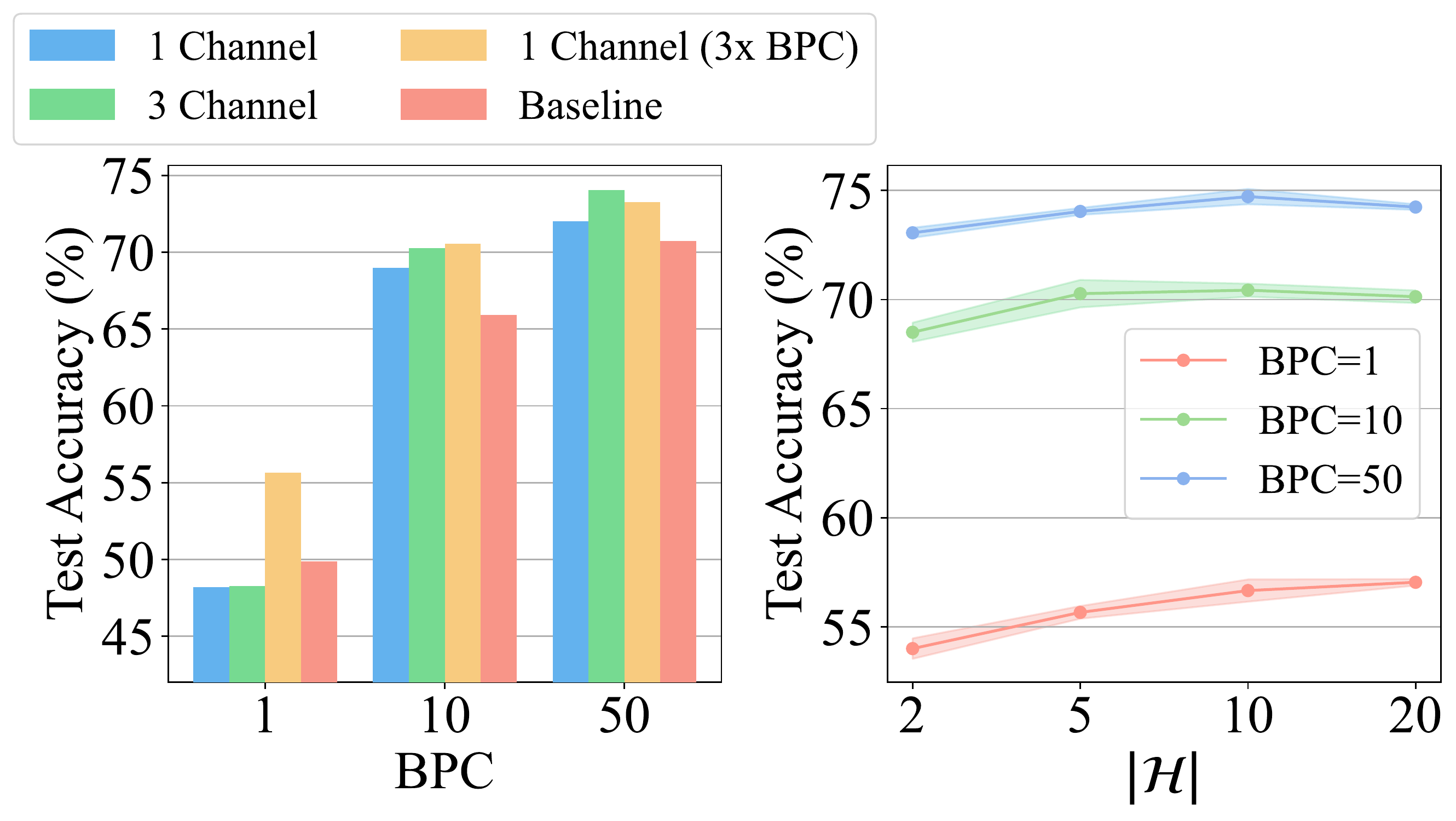}
    \caption{Study on the number of channels used by bases and the number of hallucinators.}
    \vspace{-0.4cm}
    \label{fig:channel_hallucinator}
}
\end{floatrow}
\vspace{-0.2cm}
\end{figure}

\textbf{Number of Channels Used by Basis:} 
By default, the shape of a basis is the same as that of a real image, which is generally in RGB 3-channel format. 
In fact, in Fig. \ref{fig:channel_hallucinator}(Left), we also verify that it is also possible to use single-channel basis, which can reduce the memory cost by nearly 2/3 without hurting the performance too much. 
Interestingly, if the memory cost is held the same, we can choose to use 3 times BPC to store single-channel bases, rather than 3-channel ones. 
This would yield impressive improvement on the test accuracy when BPC is small. 
Note that for baseline results, IPC is set as the corresponding BPC plus 1. 

\textbf{Number of Hallucinators:}
We study the impact of the number of hallucinators, \textit{i.e.}, $|\mathcal{H}|$, in Fig. \ref{fig:channel_hallucinator}(Right). 
We can observe that when BPC is small, including more hallucinators is helpful for the performance. 
Nevertheless, when BPC is 10 or 50, the performance would not improve with more hallucinators when $|\mathcal{H}|>10$. 
One reason is that when $|\mathcal{H}|$ is large, the sampling of hallucinators in each iteration is sparse, which makes the joint optimization of all the hallucinators more difficult. 

\textbf{Data Augmentation:} 
The similarity between our hallucinator set and data augmentation lies that both of them can contribute to generating more samples and increasing the diversity. 
However, the essential difference is that our hallucinators are optimized to encode sample-wise relationships in a dataset, while data augmentation is based on some prior and heuristic knowledge of images. 
By default, both our method and the baseline adopt the data augmentation strategy DSA~\cite{zhao2021dataset}. 
To study the relationship between the two schemes experimentally, we attempt to remove DSA from baseline and our method and report the corresponding results in Tab. \ref{tab:augment}. 
The evaluation is on CIFAR10 with 11 IPC for baseline and 10 BPC for ours. 
Through the results, we can find that (1) our method without data augmentation can also outperform the baseline method with augmentation significantly, which means that the mechanism of HaBa can benefit the dataset distillation task more with the learning of global information of a dataset in hallucinators; and (2) with data augmentation, our performance can be further improved, which indicates that HaBa and DSA work in different manners. 

\begin{figure}[!t]
\begin{floatrow}[2]
\tablebox{\caption{Study on whether all the classes should share the same set of hallucinators.}}{
    \centering
    \scriptsize
    \setlength{\tabcolsep}{2pt}
    \vspace{-5pt}
    \begin{tabular}{ccccc}
        \toprule
        & BPC & 1 & 10 & 50 \\ 
        \midrule
        \multicolumn{2}{c}{w/o Share} & 55.96$\pm$0.51 & 69.00$\pm$0.20 & 69.81$\pm$0.56 \\
        \multicolumn{2}{c}{Share} & 55.66$\pm$0.29 & 70.27$\pm$0.63 & 74.04$\pm$0.16 \\
        \multicolumn{2}{c}{Baseline (IPC=BPC)} & 45.29$\pm$0.86 & 62.77$\pm$0.56 & 71.09$\pm$0.34 \\
        \midrule
        \multicolumn{2}{c}{Share (Same Memory)} & 70.27$\pm$0.63 & 72.17$\pm$0.30 & 74.89$\pm$0.15 \\
        \multicolumn{2}{c}{Baseline (Same Memory)} & 65.92$\pm$0.62 & 68.58$\pm$0.49 & 73.55$\pm$0.48 \\
        \bottomrule
    \end{tabular}
    \label{tab:share_hallucinator}
}
\tablebox{\caption{Impact of data augmentation.}}{
    \centering
    \scriptsize
    \setlength{\tabcolsep}{2pt}
    \vspace{-5pt}
    \begin{tabular}{ccccc}
        \toprule
         & ConvNet & ResNet & VGG & AlexNet \\ 
        \midrule
        w/o aug. & 60.63$\pm$0.21 & 43.24$\pm$0.83 & 48.02$\pm$0.53 & 30.58$\pm$1.44 \\
        Baseline & 63.90$\pm$0.29 & 45.05$\pm$1.46 & 50.49$\pm$1.02 & 35.95$\pm$1.52 \\
        \midrule
        w/o aug. & 68.08$\pm$0.23 & 56.37$\pm$0.11 & 59.04$\pm$0.50 & 48.27$\pm$3.04 \\
        Ours & 69.48$\pm$0.26 & 57.97$\pm$0.88 & 60.47$\pm$0.56 & 48.96$\pm$3.00 \\
        \bottomrule
    \end{tabular}
    \label{tab:augment}
}

\end{floatrow}
\vspace{-0.7cm}
\end{figure}

\section{Conclusions, Limitations, and Future Works}

This paper proposes a novel hallucinator-basis factorization method dubbed HaBa for dataset distillation (DD). 
It uses hallucinators to encode inner relations between different samples in original datasets, which can largely improve the data efficiency of distilled results. 
To diversify the knowledge captured by different hallucinators, a pair of adversarial contrastive \xw{constraints} is further introduced. 
Extensive evaluations and comparisons on multiple benchmark datasets demonstrate that HaBa is capable of significantly improving the performance of downstream models trained on the synthetic dataset, using only 35\% cost of memory for storage. 
Moreover, it is a versatile strategy that is compatible with different configurations of DD frameworks and yields consistent improvement. 

Despite the superior performance of the proposed hallucinator-basis factorization (HaBa) scheme, there are also some potential limitations. 
On the one hand, compared with the baseline method HaBa built upon, the process of online pairwise combination between hallucinators and bases in training increases the cost of time and GPU memory slightly, although light-weight hallucinators are adopted. 
On the other hand, it may inherited the limitations of baseline methods. 
For example, when the number of images is large, further increasing the number would produce limited performance gain. 

For future works, beyond the training efficiency of HaBa, introducing class-wise relationship may also be a potential research direction. 
For example, it is probably optimal that one class shares hallucinators with some specific classes but does not share with others. 
It is also promising to explore more advance factorization for a dataset to further improve the performance. 

\section*{Acknowledgement}
This research is supported by the National Research Foundation, 
Singapore under its Medium Sized Centre 
for Advanced Robotics Technology Innovation (WBS: A-0009428-09-00).
Xinchao Wang is the corresponding author.

\medskip

{\small
\bibliographystyle{plain}
\bibliography{main}
}

\newpage
\input{check_list}

\newpage
\appendix

\begin{appendices}

In this part, we provide additional details, more results, potential limitations, and future directions of the proposed Hallucinator-Basis factorization (HaBa) for \xw{dataset disllation (DD)}. 
First, we provide more details on the pipeline of HaBa. 
Then, we conduct more experiments to demonstrate and analyze performance of our method, including results on more benchmarks with larger resolutions, as supplement to the quantitative study in the main paper. 
We also provide more qualitative results by HaBa and additional ablation studies. 
Finally, we discuss some limitations and future works of our method. 

\input{algorithm}

\section{Algorithm Details}
To better elaborate the details of the proposed HaBa for DD, we provide an algorithmic illustration for the whole pipeline in Alg. \ref{alg}, as a supplement to Sec.~3 of the main paper. 
The overall algorithm takes an original dataset as well as some hyper-parameters shown in Alg. \ref{alg} as input. 
The output is the distilled result including a set of hallucinators $\mathcal{H}$ and a set of bases $\mathcal{B}$, as defined in Eq. 1 of the main paper. 
The goal is to equip the distilled dataset with similar downstream performance to the original one. 

\section{More Results}

\begin{table}[!t]
\begin{floatrow}[1]
\tableboxf{}{
    \centering
    \scriptsize
    \setlength{\tabcolsep}{6pt}
    \vspace{-5pt}
    \begin{tabular}{cc|ccc|ccc}
    \toprule
    \multirow{3}{*}{} & Dataset & \multicolumn{3}{c|}{MNIST} & \multicolumn{3}{c}{FashionMNIST} \\
    \midrule
    & IPC & 1 & 10 & 50 & 1 & 10 & 50  \\
    & Ratio \% & 0.017 & 0.17 & 0.83 & 0.017 & 0.17 & 0.83 \\ \midrule
    \multirow{4}{*}{Coreset} & Random & 64.9$\pm$3.5 & 95.1$\pm$0.9 & 97.9$\pm$0.2 & 51.4$\pm$3.8 & 73.8$\pm$0.7 & 82.5$\pm$0.7  \\
    & Herding & 89.2$\pm$1.6 & 93.7$\pm$0.3 & 94.8$\pm$0.2 & 67.0$\pm$1.9 & 71.1$\pm$0.7 & 71.9$\pm$0.8  \\
    & K-Center & 89.3$\pm$1.5 & 84.4$\pm$1.7 & 97.4$\pm$0.3 & 66.9$\pm$1.8 & 54.7$\pm$1.5 & 68.3$\pm$0.8 \\
    & Forgetting & 35.5$\pm$5.6 & 68.1$\pm$3.3 & 88.2$\pm$1.2 & 42.0$\pm$5.5 & 53.9$\pm$2.0 & 55.0$\pm$1.1 \\
    \midrule
    \multirow{8}{*}{Distillation} & DD~\cite{wang2018dataset} & - & 79.5$\pm$8.1 & - & - & - & -  \\
    & LD~\cite{bohdal2020flexible} & 60.9$\pm$3.2 & 87.3$\pm$0.7 & 93.3$\pm$0.3 & - & - & -  \\
    & DC~\cite{zhao2020dataset} & 91.7$\pm$0.5 & 97.4$\pm$0.2 & 98.8$\pm$0.2 & 70.5$\pm$0.6 & 82.3$\pm$0.4 & 83.6$\pm$0.4  \\
    & DSA~\cite{zhao2021dataset} & 88.7$\pm$0.6 & \textbf{97.8$\pm$0.1} & \textbf{99.2$\pm$0.1} & 70.6$\pm$0.6 & 84.6$\pm$0.3 & 88.7$\pm$0.2  \\
    & DM~\cite{zhao2021distribution} & 89.7$\pm$0.6 & 97.5$\pm$0.1 & 98.6$\pm$0.1 & - & - & -  \\
    & CAFE~\cite{wang2022cafe} & \textbf{93.1$\pm$0.3} & 97.2$\pm$0.2 & 98.6$\pm$0.2 & 77.1$\pm$0.9 & 83.0$\pm$0.4 & 84.8$\pm$0.4  \\
    & CAFE+DSA~\cite{wang2022cafe} & 90.8$\pm$0.5 & 97.5$\pm$0.1 & 98.9$\pm$0.2 & 73.7$\pm$0.7 & 83.0$\pm$0.3 & 88.2$\pm$0.3  \\
    & MTT~\cite{cazenavette2022dataset} & 88.7$\pm$1.0 & 96.6$\pm$0.4 & 98.1$\pm$0.1 & 75.7$\pm$1.5 & 88.4$\pm$0.4 & 90.0$\pm$0.1 \\
    \midrule
    \multirow{3}{*}{Factorization} & BPC & 1 & 9 & 49 & 1 & 9 & 49 \\
    & Ratio \% & 0.034 & 0.17 & 0.83 & 0.034 & 0.17 & 0.83 \\
    & HaBa & 92.4$\pm$0.4 & 97.4$\pm$0.2 & 98.1$\pm$0.1 & \textbf{80.9$\pm$0.7} & \textbf{88.6$\pm$0.2} & \textbf{90.3$\pm$0.1} \\
    \midrule
    \multicolumn{2}{c|}{Whole Dataset} & \multicolumn{3}{c|}{99.6$\pm$0.0} & \multicolumn{3}{c}{93.5$\pm$0.1} \\
    \bottomrule
    \end{tabular}
    \caption{The performance (test accuracy \%) comparison with state-of-the-art methods on MNIST and FashionMNIST datasets. IPC: Number of Images Per Class; BPC: Number of Bases Per Class; Ratio~(\%): the ratio of distilled images to the whole training set.}
\label{tab:supp_compare_sota}
}
\end{floatrow}
\end{table}

\textbf{Low-Resolution Data:} 
We provide results on the more-common benchmark datasets in DD in Tab. \ref{tab:supp_compare_sota}: MNIST~\cite{lecun1998gradient} and FashionMNIST~\cite{xiao2017fashion}. 
Both datasets contain 60,000 images for training and 10,000 images for testing in 10 classes. 
The images are under $28\times28$ resolution with 1 channel. 
We build our HaBa on MTT~\cite{cazenavette2022dataset} in this part. 
Although the performances of DD on these two dataset seem to be saturated, our method may still yield consistent improvement over the baseline, especially when the ratio of distilled images to the whole training set is small. 

\textbf{ImageNet Subsets:}
We also evaluate the proposed scheme on the more-challenging settings of ImageNet~\cite{deng2009imagenet} subsets. 
We follow the baseline MTT~\cite{cazenavette2022dataset}
for the divisions of subsets.
The 6 subsets include ImageFruit, ImageMeow, ImageNette, ImageSquawk, ImageWoof, and ImageYellow. 
Each subset contains over 10,000 images, and we resize all the images to $128\times128$ resolution following the original setting. 
We use ConvNet with 5 \texttt{Conv-InstanceNorm-ReLU-AvgPool} layers for training.
For testing, in addition to the same structure of ConvNet, 
we also evaluate the results under 3 other architectures: ResNet, VGG, and AlexNet. 
To ensure the same number of parameters used for the distilled datasets, 
we set the number of images per class used by the baseline 
as the number of bases per class used by HaBa plus 1, 
\textit{i.e.}, 2 IPC v.s. 1 BPC and 11 IPC v.s. 10 BPC. 
Other settings follow the same configuration in the main paper. 

The test performances of models trained by the distilled datasets are shown in Tab. \ref{tab:supp_cross_architecture}. 
We can observe that HaBa outperforms the baseline in almost all cases except several experiments when IPC and BPC are small and the architectures of training and testing are the same. 
Notably, in all the cross-architecture generalization settings, HaBa achieves superior performance over the baseline, which further demonstrates the improvement of data efficiency introduced by the factorization and online pair-wise combination. 

\begin{table}[!t]
\begin{floatrow}[1]
\tableboxf{}{
    \centering
    \scriptsize
    \setlength{\tabcolsep}{2pt}
    \vspace{-5pt}
    \begin{tabular}{c|c|cc|cc|cc|cc}
    \toprule
    \multirow{3}{*}{} & Method & \multicolumn{2}{c|}{ConvNet} & \multicolumn{2}{c|}{ResNet} & \multicolumn{2}{c|}{VGG} & \multicolumn{2}{c}{AlexNet} \\
    \midrule
    & IPC & 2 & 11 & 2 & 11 & 2 & 11 & 2 & 11 \\
    & BPC & 1 & 10 & 1 & 10 & 1 & 10 & 1 & 10 \\ 
    \midrule
    \multirow{3}{*}{ImageFruit} & Baseline & 31.76$\pm$1.64 & 40.12$\pm$1.87 & 24.36$\pm$2.20 & 31.24$\pm$1.71 & 30.20$\pm$1.43 & 42.52$\pm$1.16 & 27.92$\pm$1.84 & 29.88$\pm$1.60 \\
    & w. HaBa & 34.68$\pm$1.13 & 42.52$\pm$1.56 & 26.60$\pm$2.48 & 33.08$\pm$1.02 & 31.92$\pm$1.91 & 45.12$\pm$1.18 & 28.16$\pm$1.29 & 32.84$\pm$1.69 \\
    & Gain & +2.92 & +2.40 & +2.24 & +1.84 & +1.72 & +2.60 & +0.24 & +2.96 \\
    \midrule
    \multirow{3}{*}{ImageMeow} & Baseline & 35.28$\pm$2.23 & 41.00$\pm$1.45 & 17.64$\pm$1.51 & 19.64$\pm$0.93 & 31.52$\pm$1.27 & 39.44$\pm$1.23 & 21.04$\pm$1.64 & 22.04$\pm$1.72 \\
    & w. HaBa & 36.92$\pm$0.93 & 42.92$\pm$0.86 & 25.44$\pm$1.02 & 26.28$\pm$2.61 & 35.00$\pm$0.76 & 47.68$\pm$0.57 & 23.76$\pm$2.06 & 24.04$\pm$1.94 \\
    & Gain & +1.64 & +1.92 & +7.80 & +6.64 & +3.48 & +8.24 & +2.72 & +2.00 \\
    \midrule
    \multirow{3}{*}{ImageNette} & Baseline & 55.16$\pm$1.08 & 63.88$\pm$0.48 & 25.52$\pm$1.31 & 42.80$\pm$1.49 & 47.48$\pm$1.67 & 62.80$\pm$1.59 & 30.96$\pm$0.97 & 34.60$\pm$2.95 \\
    & w. HaBa & 51.92$\pm$1.65 & 64.72$\pm$1.60 & 28.88$\pm$2.61 & 46.84$\pm$1.25 & 47.80$\pm$1.21 & 63.76$\pm$1.05 & 33.28$\pm$1.98 & 40.84$\pm$1.80 \\
    & Gain & -3.24 & +0.84 & +3.36 & +4.04 & +0.32 & +0.96 & +2.68 & +6.24 \\
    \midrule
    \multirow{3}{*}{ImageSquawk} & Baseline & 43.92$\pm$0.63 & 54.64$\pm$0.96 & 30.64$\pm$1.47 & 46.40$\pm$1.85 & 39.36$\pm$1.83 & 52.00$\pm$1.91 & 22.04$\pm$1.80 & 34.20$\pm$2.08 \\
    & w. HaBa & 41.88$\pm$1.37 & 56.80$\pm$1.04 & 31.52$\pm$2.39 & 48.92$\pm$1.77 & 39.64$\pm$1.78 & 56.88$\pm$0.84 & 23.28$\pm$0.55 & 35.00$\pm$1.72 \\
    & Gain & -2.04 & +2.16 & +0.88 & +2.52 & +0.28 & +4.88 & +1.24 & +0.80 \\
    \midrule
    \multirow{3}{*}{ImageWoof} & Baseline & 30.92$\pm$1.26 & 36.56$\pm$0.75 & 16.24$\pm$1.48 & 18.12$\pm$0.47 & 25.60$\pm$0.69 & 29.36$\pm$1.23 & 22.68$\pm$1.42 & 23.68$\pm$1.37 \\
    & w. HaBa & 32.40$\pm$0.67 & 38.60$\pm$1.26 & 20.20$\pm$1.55 & 25.20$\pm$0.95 & 27.08$\pm$1.81 & 37.44$\pm$1.08 & 24.88$\pm$1.20 & 27.72$\pm$1.12 \\
    & Gain & +1.48 & +2.04 & +3.96 & +7.08 & +1.48 & +8.08 & +2.20 & +4.04 \\
    \midrule
    \multirow{3}{*}{ImageYellow} & Baseline & 49.72$\pm$1.38 & 60.40$\pm$1.46 & 29.08$\pm$1.99 & 42.72$\pm$1.24 & 44.04$\pm$1.46 & 50.84$\pm$0.56  & 28.60$\pm$1.48 & 35.60$\pm$2.03 \\
    & w. HaBa & 50.44$\pm$1.56 & 63.00$\pm$1.61 & 36.32$\pm$0.65 & 48.48$\pm$1.55 & 47.28$\pm$1.59 & 57.24$\pm$1.01 & 29.08$\pm$1.19 & 36.44$\pm$1.21 \\
    & Gain & +0.72 & +2.60 & +7.24 & +5.76 & +3.24 & +6.40 & +0.48 & +0.84 \\
    \bottomrule
    \end{tabular}
    \vspace{-0.1cm}
    \caption{Cross-architecture performance (test accuracy \%) comparison with the baseline on various subsets of ImageNet dataset.}
    
\label{tab:supp_cross_architecture}
}
\end{floatrow}
\vspace{-0.4cm}
\end{table}


\begin{figure}[!t]
\begin{floatrow}[2]
\tablebox{\caption{\xw{Ablation studies on the depth (number of nonlinear blocks) of hallucinator.}}}{
    \centering
    \scriptsize
    \setlength{\tabcolsep}{1pt}
    \vspace{-5pt}
    \begin{tabular}{ccccc}
        \toprule
        Depth & 0 & 1 & 2 & 3 \\ 
        \midrule
        Accuracy (\%) & 68.43$\pm$0.37 & 70.27$\pm$0.63 & 71.17$\pm$0.29 & 71.55$\pm$0.27 \\
        Downstream Speed & 144.54 & 140.11 & 125.04 & 115.62 \\
        \# of Parameters & 6,144 & 6,312 & 10,963 & 16,131 \\
        \bottomrule
    \end{tabular}
    \label{tab:depth}
}
\tablebox{\caption{\xw{Ablation studies on the number of feature channels in hallucinator.}}}{
    \centering
    \scriptsize
    \setlength{\tabcolsep}{1pt}
    \vspace{-5pt}
    \begin{tabular}{ccccc}
        \toprule
        \# of Channels & 3 & 8 & 16 \\ 
        \midrule
        Accuracy & 70.27$\pm$0.63 & 70.47$\pm$0.37 & 71.28$\pm$0.35 \\
        Downstream Speed & 140.11 & 138.48 & 135.12 \\
        \# of Parameters & 6,312 & 16,827 & 33,651 \\
        \bottomrule
    \end{tabular}
    \label{tab:channel}
}
\end{floatrow}
\end{figure}

\begin{figure}[!t]
\begin{floatrow}[2]
\tablebox{\caption{\xw{Impact on sharing encoder and decoder across all hallucinators.}}}{
    \centering
    \scriptsize
    \setlength{\tabcolsep}{2pt}
    \vspace{-5pt}
    \begin{tabular}{cccc}
        \toprule
        BPC & 1 & 10 & 50 \\ 
        \midrule
        Ours & 55.66$\pm$0.29 & 70.27$\pm$0.63 & 74.04$\pm$0.16 \\
        Share Enc. \& Dec. & 55.14$\pm$0.44 & 69.47$\pm$0.09 & 72.69$\pm$0.39 \\
        Baseline & 49.89$\pm$0.95 & 65.92$\pm$0.62 & 70.73$\pm$0.52 \\
        \bottomrule
    \end{tabular}
    \label{tab:share_enc_dec}
}
\tablebox{\caption{\xw{Results of speech recognition on Mini Speech Commands.}}}{
    \centering
    \scriptsize
    \setlength{\tabcolsep}{2pt}
    \vspace{-5pt}
    \begin{tabular}{c|cccccc|c}
        \toprule
        SPC & Rand & Herd & DSA & DM & IDC & IDC w. HaBa & Whole Dataset \\ 
        \midrule
        10 & 42.6 & 56.2 & 65.0 & 69.1 & 73.3 & 74.5 & \multirow{2}{*}{93.4} \\
        20 & 57.0 & 72.9 & 74.0 & 77.2 & 83.0 & 84.3 & \\
        \bottomrule
    \end{tabular}
    \label{tab:speech}
}
\end{floatrow}
\end{figure}

\textbf{More Ablations on Hallucinators:}
\xw{
In the default setting, the encoder and decoder of hallucinators have 1 \texttt{Conv-ReLU} block and the number of feature channels is 3. 
In this part, we provide more results when we consider increasing the capacity of the hallucination networks. 
As shown in Tab. \ref{tab:depth}, we try increasing the depth of the networks by adding more nonlinear blocks. 
Although the performance can indeed be improved, it results in nonnegligible latency to downstream training speed, measured by the number of epochs per second. 
Taking both training speed and performance into consideration, we consider using only 1 nonlinear block by default, which yields best trade-off between the two factors. 
Likewise, we also try increasing the number of feature channels in halluciantors as shown in Tab. \ref{tab:channel}. 
The number of parameters is almost proportional to the number of channels. 
However, the performance gain is very limited. 
Thus, we simply take the number of channels in images, which is 3 for RGB images, as the number of feature channels in hallucinators. 
}

\begin{figure}[!t]
\begin{floatrow}[1]
\figureboxf{}{
  \centering
  \includegraphics[width=0.9\textwidth]{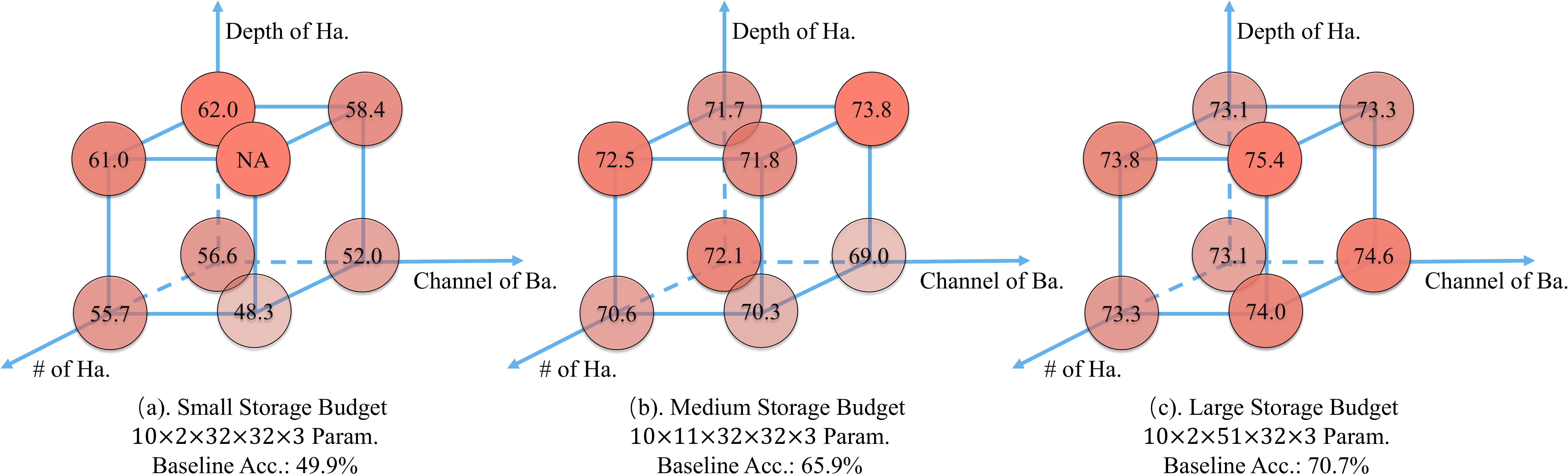}
  \caption{Exploration on the configurations of different factors in hallucinators and bases.}
  \label{fig:search}
}
\end{floatrow}
\vspace{-0.6cm}
\end{figure}

\textbf{More Insights on the Configurations of Hallucinators and Bases:} 
\xw{
As shown in the ablation studies in the main paper and the supplement, under the framework of hallucinator-basis factorization, there are many factors that affect the performance. 
Given a fixed storage budget, how to scale the bases and hallucinators is an important topic. 
Among all the factors, we empirically find that the  depth of hallucinators, the number of hallucinators, the number of  channels in each basis, and the number of bases are the most important ones, which will be studied in the following exploration. 
Here, we consider three types of storage budget: small, medium, and large, corresponding to the cost of IPC=2, 11, and 51 for the baseline method respectively. 
We consider cases of 1 and 2 convolution blocks for the depth of hallucinators, 2 and 5 for the number of hallucinators, and 1 and 3 for the number of channels in each basis. For each setting, we adjust the number of bases to fit the given budget. 
Enumerating all the configurations, there are totally 8 settings for each kind of budget. 
Their results are visualized in Fig. \ref{fig:search}. 
Based on the results, we have the following observations:
\begin{itemize}
    \item \textbf{For all the three types of budget, the best performance is achieved by using deeper hallucinators. Especially under small and medium budgets,  using depth 2 can outperform using depth 1 almost consistently.} This  can be explained by the more complex sample-wise relationship extracted  by hallucinators.
    \item In our framework, bases are expected to store sample-independent information while hallucinators are used to encode shared relationship across all the samples. \textbf{When the budget is small, using 1-channel bases can achieve significantly better results.} This is because small storage budget would more rely on increasing the number of independent data samples for a better diversity. The informativeness of each basis appears less important.
    \item \textbf{When the budget increases, the advantage of 1-channel bases mentioned before would diminish gradually. Especially under the large budgets,  3-channel bases outperform 1-channel ones consistently.} The reason is  that when the number of bases is adequate, focusing on the informativeness of each basis can produce more benefit than increasing the number.
    \item \textbf{When the budget is large, using more hallucinators can yield slightly better results}, which can probably be attributed to the further improvement on the diversity.
    \item \textbf{The larger the budget is, the less insensitive the performance is, to different configurations.} 
\end{itemize}
}
Note that the above exploration is conducted without taking the downstream training speed into consideration, which is also an important metric in the task of dataset distillation. Our opinion on the scalability is that, when downstream training overhead is not a issue, deeper hallucinators are recommended for better performance; otherwise if downstream efficiency is desired, we find that 1 nonlinear block is sufficient, since heavier hallucination networks can result in nonnegligible latency, especially when the total number of images is large.

\textbf{Sharing Encoder and Decoder across all Hallucinators:} 
\xw{
As a variant of our default case which uses different hallucination networks, it is also feasible for the halluciantors to share a common group of encoder and decoder but use different parameters $(\sigma,\mu)$ for affine transformation, which is potential to further boost the data efficiency. 
As shown in Tab. \ref{tab:share_enc_dec}, the performance becomes slightly worse. 
We conjecture that different convolution encoders and decoders may contribute to the diversity of the extracted patterns, which increases the representation ability of the hallucinator set. Moreover, since we only use 1 convolution block for encoders and decoders, the number of parameters is not so significant compared with that of a basis. 
Therefore, we consider making the whole network independent with each other for all hallucinators. 
}

\textbf{Results on Speech Domain:} 
\xw{
To validate the versatility of the proposed hallucinator-basis factorization solution, we further conduct experiments on the speech domain using Mini Speech Commands~\cite{warden2018speech}, which contains 8,000 audio clips for 8 command classes. We adopt IDC~\cite{kim2022dataset} as the baseline and all the protocols for comparisons follow the official settings. 
We compare our method with the coreset selection based Random and Herding, DSA~\cite{zhao2021dataset}, DM~\cite{zhao2021distribution}, and the IDC baseline~\cite{kim2022dataset}. 
The results in Tab. \ref{tab:speech} shows that our method can produce consistent improvement on the downstream test accuracy, which further reflects the generality of our method for different modalities. 
Here, SPC denotes the number of speech spectrograms per class. 
}

\begin{figure}[!t]
\begin{floatrow}[2]
\figureboxl{\caption{Generalization performance on images with different corrupted levels.}}{
    \includegraphics[width=0.99\linewidth]{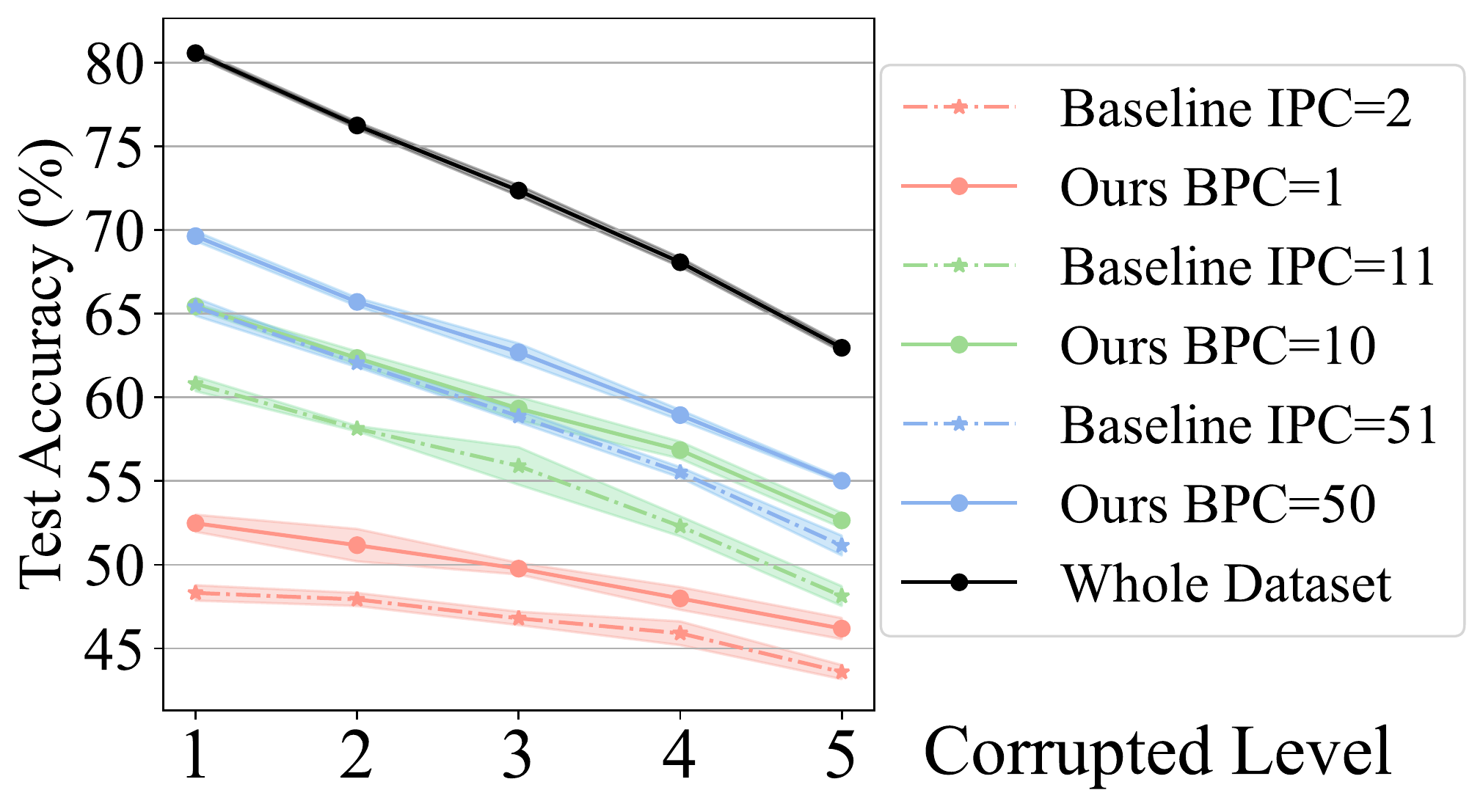}
    \label{fig:corrupt}
    \vspace{-0.4cm}
}
\tableboxl{\caption{\xw{List of hyper-parameters.}}}{
    \centering
    \scriptsize
    \setlength{\tabcolsep}{2pt}
    \vspace{-5pt}
    \begin{tabular}{ccc}
        \toprule
        Hyper-Parameter & Notation & Value \\ 
        \midrule
        Height of Basis & $h'$ & Height of Image $h$ \\
        Width of Basis & $w'$ & Width of Image $w$ \\
        Channel of Basis & $c'$ & Channel of Image $c$ \\
        Channel of Hallucinator & $c''$ & Channel of Image $c$ \\
        Depth of Hallucinator & - & 1 \\
        Learning Rate of Feature Extractor & $\eta_F$ & 0.001 \\
        Weight of $\mathcal{L}_{con.}$ & $\lambda_{con.}$ & 0.1 \\
        Weight of $\mathcal{L}_{task}$ & $\lambda_{task}$ & 1 \\
        Weight of $\mathcal{L}_{DD}$ & $\lambda_{DD}$  & 1 \\
        Weight of $\mathcal{L}_{cos.}$ & $\lambda_{cos.}$ & 0.1 \\
        \bottomrule
    \end{tabular}
    \label{tab:hyper}
}
\end{floatrow}
\vspace{-0.6cm}
\end{figure}

\textbf{Robustness to Corruption:} 
We further examine the generalization performance of our method and the baseline one on CIFAR10-C~\cite{hendrycks2019robustness}, the corrupted version of CIFAR10 dataset with 19 different types of corruption. 
There are five corrupted levels from 1 (mildest) to 5 (severest) and we report the mean test accuracy across 19 domains on different levels. 
\xw{Since the proposed method can increase the accuracy and alleviate the under-fitting problem on the original domain, which is one dominant component of cross-domain generalization~\cite{ben2010theory}}, it can also demonstrate superior robustness in all corrupted data as demonstrated in Fig. \ref{fig:corrupt}. 
Also, the gap between performance using distilled dataset and original dataset becomes smaller with the increase of corrupted level, which suggests that our method improves the domain generalization ability potentially, thanks to the diverse training data composed of hallucinators and bases. 

\textbf{List of Hyper-Parameters:} 
\xw{
In Tab. \ref{tab:hyper}, we provide a clear view of the hyper-parameters used in this paper. 
All the experiments follows these settings if not specified. 
The performance of our method is insensitive to the values of these hyper-parameters as analyzed in the ablation studies in both the main paper and the appendix. 
Other hyper-parameters not listed come from the adopted baseline methods and we follow their original settings. 
}

\begin{figure}[!t]
\begin{floatrow}[1]
\figureboxf{}{
  \centering
  \begin{subfigure}{0.245\linewidth}
  \includegraphics[width=\linewidth]{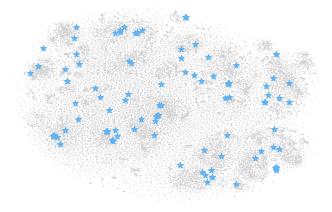}
  \vspace{-0.7cm}
  \caption{MTT on CIFAR10}
  \end{subfigure}
  \begin{subfigure}{0.245\linewidth}
  \includegraphics[width=\linewidth]{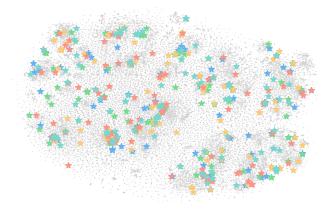}
  \vspace{-0.7cm}
  \caption{HaBa on CIFAR10}
  \end{subfigure}
  \begin{subfigure}{0.245\linewidth}
  \includegraphics[width=\linewidth]{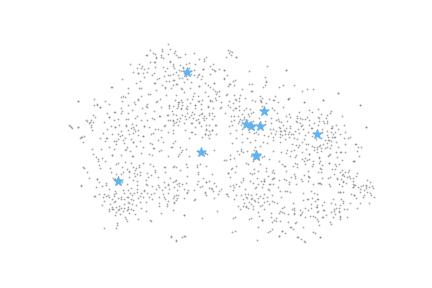}
  \vspace{-0.7cm}
  \caption{MTT on ImageSquawk}
  \end{subfigure}
  \begin{subfigure}{0.245\linewidth}
  \includegraphics[width=\linewidth]{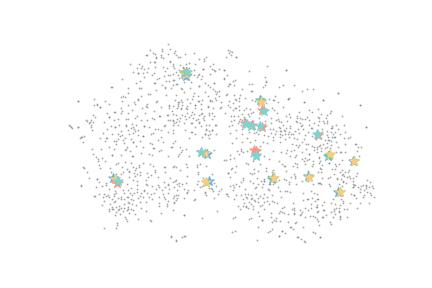}
  \vspace{-0.7cm}
  \caption{HaBa on ImageSquawk}
  \end{subfigure}
  \vspace{-0.3cm}
  \caption{TSNE visualization of results by our HaBa and baseline MTT on CIFAR10 and ImageSquawk datasets. Markers with different colors in our results denote images generated by different hallucinators. Gray dots denote real images.}
  \label{fig:tsne}
  }
  \end{floatrow}
  \vspace{-0.3cm}
\end{figure}

\textbf{TSNE Visualizations:} 
To provide a better understanding on why the HaBa factorization can help on data efficiency in dataset distillation, we adopt TSNE~\cite{van2008visualizing} to visualize the features before the last linear layer of a teacher model trained on the original datasets. 
In Fig. \ref{fig:tsne}, we plot features of both original images and the distilled ones. 
The results reveal that datasets restored from our hallucinators and bases can describe the original data distribution more finely, which means that the original datasets can be represented with the distilled ones with less information loss. 
Given that the total numbers of parameters used for storing distilled datasets are the same, our method can improve the data efficiency significantly. 

\begin{figure}[!t]
\begin{floatrow}[1]
\figureboxf{}{
  \centering
  \begin{subfigure}{0.32\linewidth}
  \includegraphics[width=\linewidth]{Figure/basis.pdf}
  \vspace{-0.7cm}
  \caption{Bases}
  \end{subfigure}
  \begin{subfigure}{0.32\linewidth}
  \includegraphics[width=\linewidth]{Figure/style_1.pdf}
  \vspace{-0.7cm}
  \caption{Images by $H_1$}
  \end{subfigure}
  \begin{subfigure}{0.32\linewidth}
  \includegraphics[width=\linewidth]{Figure/style_3.pdf}
  \vspace{-0.7cm}
  \caption{Images by $H_2$}
  \end{subfigure}
  \begin{subfigure}{0.32\linewidth}
  \includegraphics[width=\linewidth]{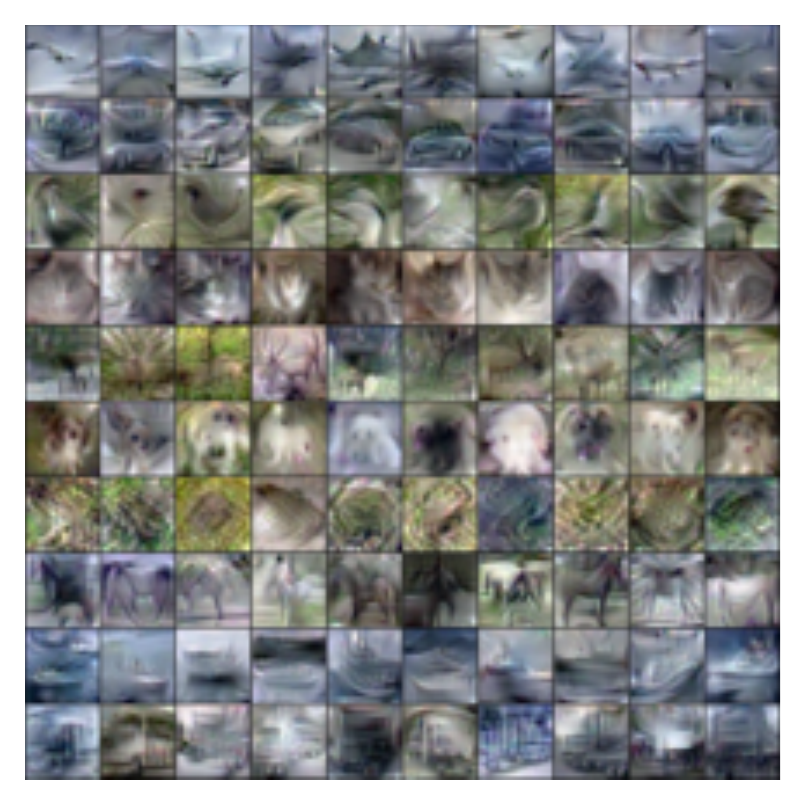}
  \vspace{-0.7cm}
  \caption{Images by $H_3$}
  \end{subfigure}
  \begin{subfigure}{0.32\linewidth}
  \includegraphics[width=\linewidth]{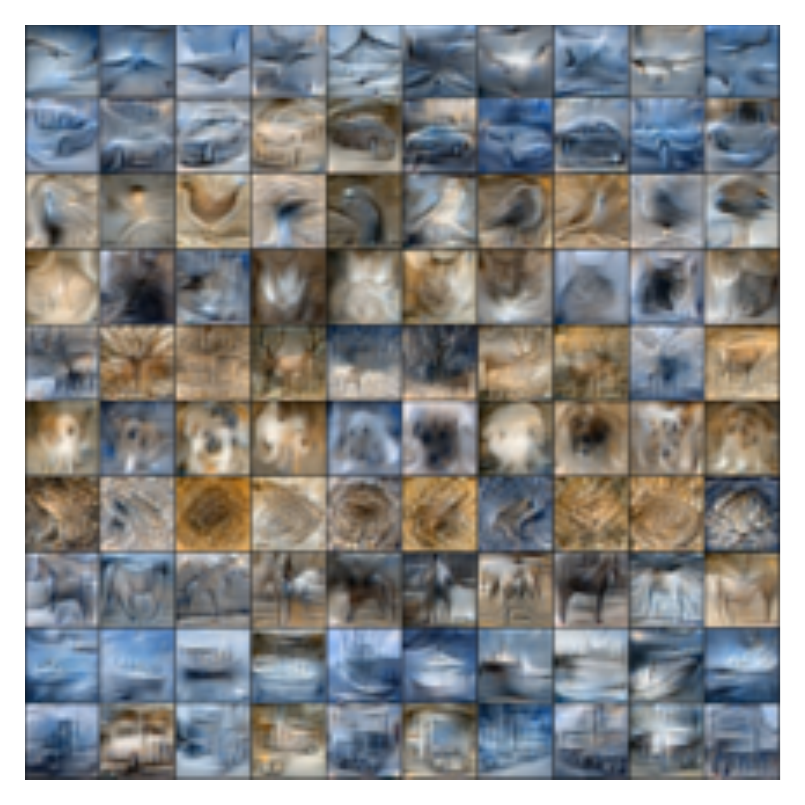}
  \vspace{-0.7cm}
  \caption{Images by $H_4$}
  \end{subfigure}
  \begin{subfigure}{0.32\linewidth}
  \includegraphics[width=\linewidth]{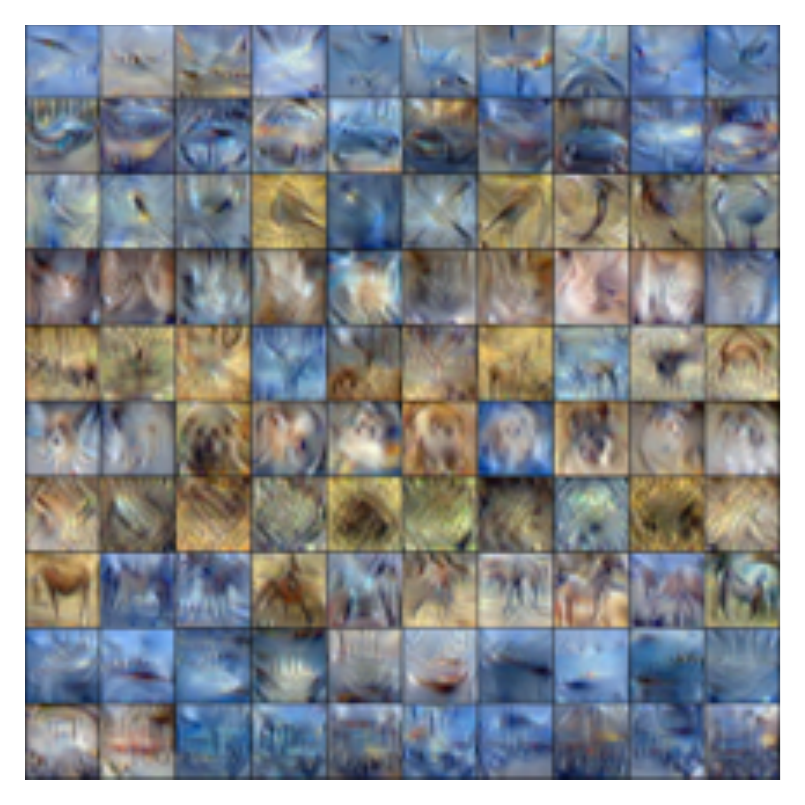}
  \vspace{-0.7cm}
  \caption{Images by $H_5$}
  \end{subfigure}
  \vspace{-0.3cm}
  \caption{Visualization of factorized results by our HaBa on CIFAR10. Please zoom-in for better visualization.}
  \label{fig:supp_visualization}
  }
  \end{floatrow}
  \vspace{-0.5cm}
\end{figure}

\begin{figure}[!t]
\begin{floatrow}[1]
\figureboxf{}{
  \centering
  \begin{subfigure}{\linewidth}
  \includegraphics[width=\linewidth]{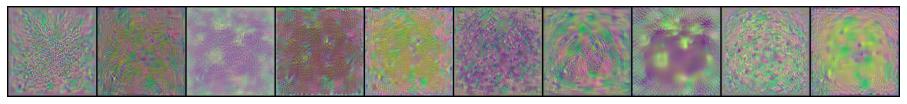}
  \caption{Bases on ImageFruit.}
  \end{subfigure}
  \begin{subfigure}{\linewidth}
  \includegraphics[width=\linewidth]{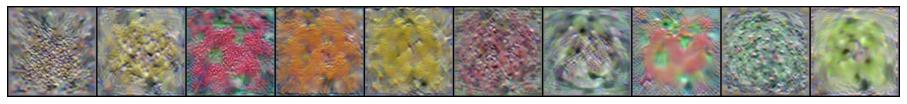}
  \caption{Images by $H_1$ on ImageFruit.}
  \end{subfigure}
  \begin{subfigure}{\linewidth}
  \includegraphics[width=\linewidth]{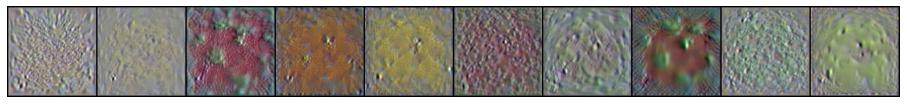}
  \caption{Images by $H_2$ on ImageFruit.}
  \end{subfigure}
  
  \vspace{0.7cm}
  
  \begin{subfigure}{\linewidth}
  \includegraphics[width=\linewidth]{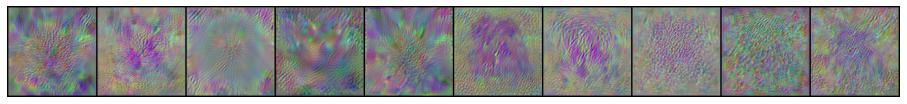}
  \caption{Bases on ImageMeow.}
  \end{subfigure}
  \begin{subfigure}{\linewidth}
  \includegraphics[width=\linewidth]{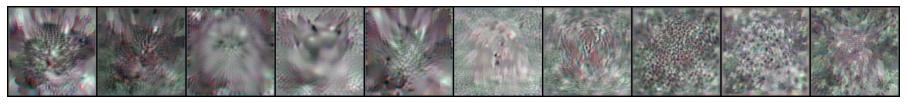}
  \caption{Images by $H_1$ on ImageMeow.}
  \end{subfigure}
  \begin{subfigure}{\linewidth}
  \includegraphics[width=\linewidth]{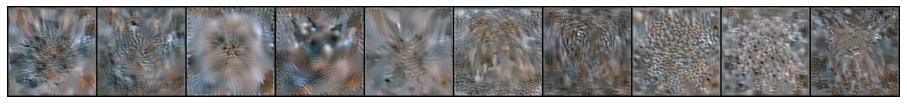}
  \caption{Images by $H_2$ on ImageMeow.}
  \end{subfigure}
  \begin{subfigure}{\linewidth}
  
  \vspace{0.7cm}
  
  \includegraphics[width=\linewidth]{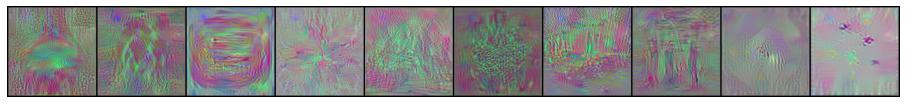}
  \caption{Bases on ImageNette.}
  \end{subfigure}
  \begin{subfigure}{\linewidth}
  \includegraphics[width=\linewidth]{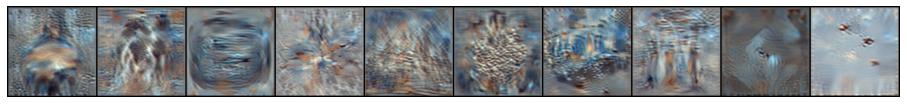}
  \caption{Images by $H_1$ on ImageNette.}
  \end{subfigure}
  \begin{subfigure}{\linewidth}
  \includegraphics[width=\linewidth]{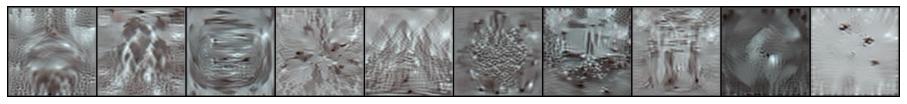}
  \caption{Images by $H_2$ on ImageNette.}
  \end{subfigure}
  
  \caption{Visualization of factorized results by our HaBa on ImageNet subsets.}
  \label{fig:imagenet_visualization}
  }
  \end{floatrow}
  \vspace{-0.5cm}
\end{figure}

\begin{figure}[!t]
\begin{floatrow}[1]
\figureboxf{}{
  \centering
  \begin{subfigure}{\linewidth}
  \includegraphics[width=\linewidth]{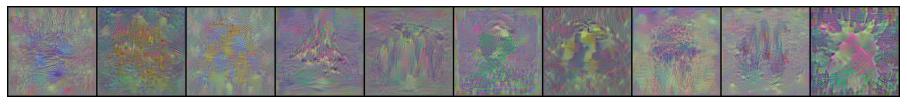}
  \caption{Bases on ImageSquawk.}
  \end{subfigure}
  \begin{subfigure}{\linewidth}
  \includegraphics[width=\linewidth]{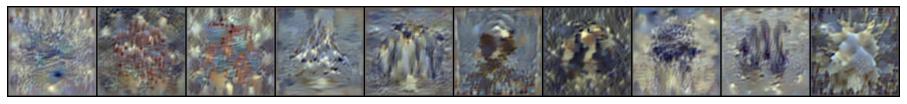}
  \caption{Images by $H_1$ on ImageSquawk.}
  \end{subfigure}
  \begin{subfigure}{\linewidth}
  \includegraphics[width=\linewidth]{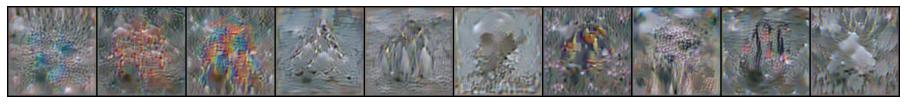}
  \caption{Images by $H_2$ on ImageSquawk.}
  \end{subfigure}
  
  \vspace{0.7cm}
  
  \begin{subfigure}{\linewidth}
  \includegraphics[width=\linewidth]{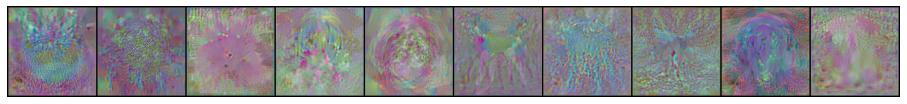}
  \caption{Bases on ImageWoof.}
  \end{subfigure}
  \begin{subfigure}{\linewidth}
  \includegraphics[width=\linewidth]{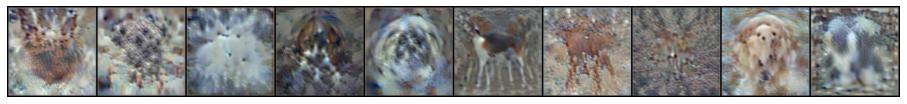}
  \caption{Images by $H_1$ on ImageWoof.}
  \end{subfigure}
  \begin{subfigure}{\linewidth}
  \includegraphics[width=\linewidth]{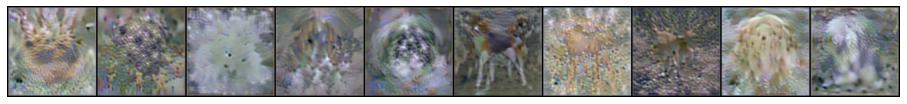}
  \caption{Images by $H_2$ on ImageWoof.}
  \end{subfigure}
  \begin{subfigure}{\linewidth}
  
  \vspace{0.7cm}
  
  \includegraphics[width=\linewidth]{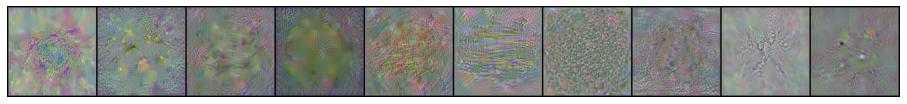}
  \caption{Bases on ImageYellow.}
  \end{subfigure}
  \begin{subfigure}{\linewidth}
  \includegraphics[width=\linewidth]{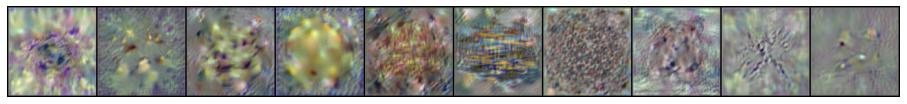}
  \caption{Images by $H_1$ on ImageYellow.}
  \end{subfigure}
  \begin{subfigure}{\linewidth}
  \includegraphics[width=\linewidth]{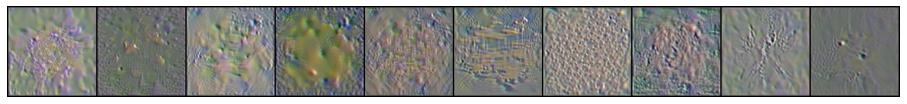}
  \caption{Images by $H_2$ on ImageYellow.}
  \end{subfigure}
  
  \caption{Visualization of factorized results by our HaBa on ImageNet subsets (Cont.).}
  \label{fig:imagenet_visualization_cont}
  }
  \end{floatrow}
  \vspace{-0.5cm}
\end{figure}

\textbf{Visualizations of Factorized Results:} 
We first provide the full results of HaBa factorization on CIFAR10 dataset with 5 hallucinators and 10 BPC in Fig. \ref{fig:supp_visualization}, as a supplement to Fig. 4 in the main paper. 
We also provide the distilled results on datasets with larger resolutions in Fig. \ref{fig:imagenet_visualization} and \ref{fig:imagenet_visualization_cont} for the above 6 ImageNet subsets. 
Here, we use 1 BPC and 2 hallucinators for visualization. 
Through these results, we can find that bases in our scheme mainly define the basic contents, while different hallucinators may transform each basis to different appearances and styles. 
Such difference is encouraged to be as large as possible to diversify the distilled data and thus improve data efficiency during the end-to-end training pipeline of DD. 

\clearpage

\end{appendices}

\end{document}

%% file: check_list.tex
\section*{Checklist}

\begin{enumerate}

\item For all authors...
\begin{enumerate}
  \item Do the main claims made in the abstract and introduction accurately reflect the paper's contributions and scope?
    \answerYes{}
  \item Did you describe the limitations of your work?
    \answerYes{Please refer to the supplement.}
  \item Did you discuss any potential negative societal impacts of your work?
    \answerNA{}
  \item Have you read the ethics review guidelines and ensured that your paper conforms to them?
    \answerYes{}
\end{enumerate}

\item If you are including theoretical results...
\begin{enumerate}
  \item Did you state the full set of assumptions of all theoretical results?
    \answerNA{}
        \item Did you include complete proofs of all theoretical results?
    \answerNA{}
\end{enumerate}

\item If you ran experiments...
\begin{enumerate}
  \item Did you include the code, data, and instructions needed to reproduce the main experimental results (either in the supplemental material or as a URL)?
    \answerYes{Please refer to the supplement.}
  \item Did you specify all the training details (e.g., data splits, hyperparameters, how they were chosen)?
    \answerYes{Please refer to Sec. \ref{sec:4-1}.}
        \item Did you report error bars (e.g., with respect to the random seed after running experiments multiple times)?
    \answerYes{We run all the experiments 5 times and report the mean and standard deviation of the performance.}
        \item Did you include the total amount of compute and the type of resources used (e.g., type of GPUs, internal cluster, or cloud provider)?
    \answerYes{Please refer to Sec. \ref{sec:4-1}.}
\end{enumerate}

\item If you are using existing assets (e.g., code, data, models) or curating/releasing new assets...
\begin{enumerate}
  \item If your work uses existing assets, did you cite the creators?
    \answerYes{}
  \item Did you mention the license of the assets?
    \answerYes{}
  \item Did you include any new assets either in the supplemental material or as a URL?
    \answerYes{}
  \item Did you discuss whether and how consent was obtained from people whose data you're using/curating?
    \answerNA{}
  \item Did you discuss whether the data you are using/curating contains personally identifiable information or offensive content?
    \answerNA{}
\end{enumerate}

\item If you used crowdsourcing or conducted research with human subjects...
\begin{enumerate}
  \item Did you include the full text of instructions given to participants and screenshots, if applicable?
    \answerNA{}
  \item Did you describe any potential participant risks, with links to Institutional Review Board (IRB) approvals, if applicable?
    \answerNA{}
  \item Did you include the estimated hourly wage paid to participants and the total amount spent on participant compensation?
    \answerNA{}
\end{enumerate}

\end{enumerate}

%% file: algorithm.tex
\begin{algorithm}
  \caption{Hallucinator-Basis Factorization (HaBa) for Dataset Distillation.}
  \begin{algorithmic}[1]
    \Require
        $\mathcal{T}$: original dataset; 
        $|\mathcal{H}|$: total number of hallucinators; 
        $|\mathcal{B}|$: total number of bases; 
        $\eta_H$: learning rate of hallucinators;
        $\eta_B$: learning rate of bases;
        $\eta_F$: learning rate of feature extractor.
    \Ensure
        $\mathcal{H}$: a set of hallucinators; 
        $\mathcal{B}$: a set of bases; 
    \State Randomly initialize hallucinators $\mathcal{H}=\{H_{\theta_j}\}_{j=1}^{|\mathcal{H}|}$, bases $\mathcal{B}=\{(\hat{x}_i,\hat{y}_i)\}_{i=1}^{|\mathcal{B}|}$, and parameters $\psi$ of the feature extractor $F$;
    \While{not done}
        \State $\mathcal{H}'\leftarrow$ a random batch of hallucinators from $\mathcal{H}$;
        \State $\mathcal{B}'\leftarrow$ a random batch of bases from $\mathcal{B}$;
        \For{each $1\leq i\leq |\mathcal{B}'|$}
            \For{each $1\leq j\leq |\mathcal{H}'|$}
                \State $\tilde{x}_{ij}=H_{\theta_j}(\hat{x}_i)$;
            \EndFor
        \EndFor
        \State Compute $\mathcal{L}_{\mathcal{S}}$ using Eq. 6 of the main paper;
        \State Update $\mathcal{H}$: $\theta_i\leftarrow\theta_i-\eta_H\nabla_{\theta_i}\mathcal{L}_{\mathcal{S}}$ for $1\leq j\leq |\mathcal{H}'|$;
        \State Update $\mathcal{B}$: $\hat{x}_i\leftarrow\hat{x}_i-\eta_B\nabla_{\hat{x}_i}\mathcal{L}_{\mathcal{S}}$ and $\hat{y}_i\leftarrow\hat{y}_i-\eta_B\nabla_{\hat{y}_i}\mathcal{L}_{\mathcal{S}}$ (optional) for $1\leq i\leq |\mathcal{B}'|$;
        \State Compute $\mathcal{L}_F$ using Eq. 4 of the main paper;
        \State Update $F$: $\psi\leftarrow\psi-\eta_F\nabla_{\psi}\mathcal{L}_F$;
    \EndWhile
  \end{algorithmic}
  \label{alg}
\end{algorithm}